\documentclass[preprint,twocolumn,10pt]{jmlr}

\usepackage{booktabs}
\usepackage{siunitx}

\usepackage[switch]{lineno}

\usepackage{amsmath,amsfonts,bm}

\def\eqref#1{equation~\ref{#1}}

\def\1{\bm{1}}

\DeclareMathAlphabet{\mathsfit}{\encodingdefault}{\sfdefault}{m}{sl}
\SetMathAlphabet{\mathsfit}{bold}{\encodingdefault}{\sfdefault}{bx}{n}

\DeclareMathOperator*{\argmin}{arg\,min}

 \DeclareMathOperator{\X}{\mathbf{X}}
\usepackage{dsfont}
\usepackage{wrapfig}
\usepackage{adjustbox}
\usepackage{graphicx}
\usepackage{caption}
\usepackage{multirow}
\usepackage{booktabs}
\usepackage{colortbl}

\newcommand{\equal}[1]{{\hypersetup{linkcolor=black}\thanks{#1}}}

\theorembodyfont{\upshape}
\theoremheaderfont{\scshape}
\theorempostheader{:}
\theoremsep{\newline}

\newcommand{\update}[1]{{\color{black} #1}}

\jmlryear{2024}

\jmlrworkshop{Under Review} 
\title[Dynamic Survival Analysis for Early Event Prediction]{Dynamic Survival Analysis for Early Event Prediction}

\author{\Name{Hugo Yèche}\equal{First co-authors}
\Email{hyeche@ethz.ch}\\
\addr ETH Zürich, Department of Computer Science, Switzerland\\
\Name{Manuel Burger}\footnotemark[1] \Email{manuel.burger@ethz.ch}\\
\addr ETH Zürich, Department of Computer Science, Switzerland\\
\Name{Dinara Veshchezerova} \Email{dinara.veshchezerova@ethz.ch}\\
\addr ETH Zürich, Department of Computer Science, Switzerland\\
\Name{Gunnar Rätsch}
\Email{raetsch@inf.ethz.ch}\\
\addr ETH Zürich, Department of Computer Science, Switzerland\\
}

\begin{document}

\maketitle

\begin{abstract}
This study advances Early Event Prediction (EEP) in healthcare through Dynamic Survival Analysis (DSA), offering a novel approach by integrating risk localization into alarm policies to enhance clinical event metrics. By adapting and evaluating DSA models against traditional EEP benchmarks, our research demonstrates their ability to match EEP models on a time-step level and significantly improve event-level metrics through a new alarm prioritization scheme (up to 11\% AuPRC difference). This approach represents a significant step forward in predictive healthcare, providing a more nuanced and actionable framework for early event prediction and management.
\end{abstract}

\paragraph*{Data and Code Availability}

This paper uses the MIMIC-III dataset (version 1.4)~\citep{mimic-iii-johnson-nature16} available
on PhysioNet~\citep{mimic-iii-johnson-physionet16} and the HiRID dataset (version 1.1.1)~\citep{faltys2021hirid}
also available on Physionet~\citep{hirid-faltys-physionet21}. We provide a code repository\footnote{\url{https://anonymous.4open.science/r/dsa-for-eep}}.

\paragraph*{Institutional Review Board (IRB)}
This research does not require IRB approval in the country it was performed in.

\section{Introduction}\label{introduction}

\begin{figure}
    \centering
    \includegraphics[width=\linewidth]{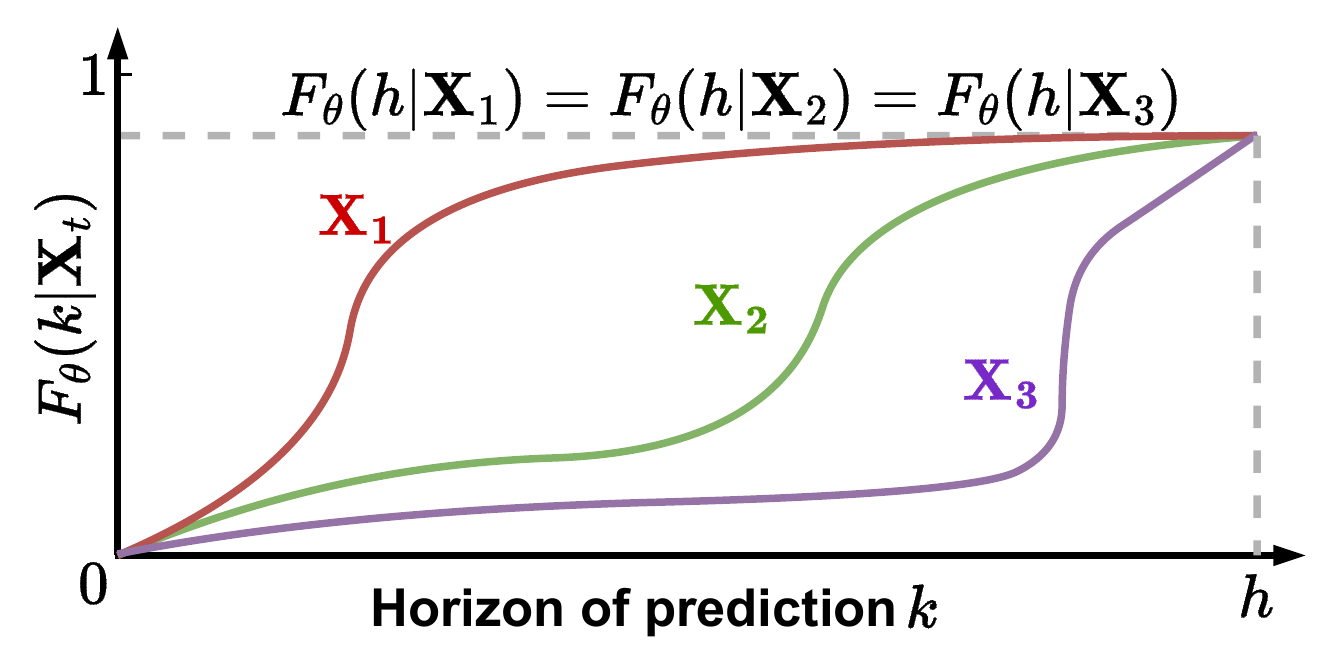}
    \caption{Localization problem when estimating failure function $F_\theta$. If only provided with an estimate at $h$, as in EEP modeling, alarms would be raised similarly for $\X_1, \X_2$ and $\X_3$. If on the other hand provided with earlier estimates like depicted, as in DSA modeling, an event for $\X_1$ is more probable to happen earlier than $\X_2$ and $\X_3$. Because of the imminence of the risk, we argue that given that information, $\X_1$ should have a higher risk score for the alarm policy. We further develop this idea in Section~\ref{sec:alarm-policy} }
    \label{fig:mot-local}
\end{figure}

Early event prediction (EEP) on time series is concerned with determining whether an event will occur within a fixed time horizon. It is highly relevant to a wide range of monitoring applications in fields such as environment~\citep{di2016potential} or healthcare~\citep{sutton2020overview}. Using machine learning for EEP has gained particular interest in Intensive Care Unit (ICU) patient monitoring~\citep{harutyunyan2019multitask,hyland2020,yeche2021,vandewaterYetAnotherICUBenchmark2023}, where large quantities of medical data are collected automatically. Existing works train such models through maximum likelihood estimation (MLE) of the cumulative failure function for a fixed horizon. However, to be usable by clinicians at an event scale, one needs to design an alarm mechanism leveraging the time-step level failure estimates. If existing works have proposed various ways of evaluating EEP models at event scale~\citep{tomavsev2019clinically,hyland2020}, the design of the alarm policy based on time-step prediction has been overlooked. Current approaches~\citep{tomavsev2019clinically,hyland2020,respiratory-hueser-2024,kidney-lyu-2024} rely on a simple fixed threshold mechanism on the time-step prediction to raise alarms at the event scale. One limitation to more advanced policies is that due to their cumulative nature current EEP models do not provide information concerning the imminence of the risk within the considered horizon as depicted in Figure~\ref{fig:mot-local}.

Parallelly, in statistics, survival analysis (SA), also known as time-to-event analysis, considers the highly related problem of predicting the \textit{exact} time of a future event given a set of covariates. With deep learning emergence, the field has also recently pivoted to discrete-time methods using neural networks~\citep{tutz2016modeling,gensheimer2019scalable,kvamme2019time, lee2018deephit,ren2019deep} to fit hazard or probability mass functions (PMF). The extension of SA to longitudinal covariates, namely dynamic survival analysis, also gained popularity in the deep learning field~\citep{lee2019dynamic, jarrett2019dynamic,damera2022intervene,maystre2022temporally}. As opposed to models trained to maximize EEP likelihood, DSA models estimate the event PMF at any horizon. Thus, in theory, such a model can also provide an estimate of the cumulative failure function as required in EEP tasks while additionally providing a decomposition of where such a risk lies within the considered horizon. 

In this work, we study the usage of deep learning models trained with a DSA likelihood for EEP tasks to design more advanced alarm policies. Our contribution can be summarized as follows: (\textbf{i}) We formalize and propose how to train and use DSA models to match EEP models' timestep-level performance on three established benchmarks \update{(\textbf{ii}) To this end, we propose  \textit{survTLS}, a non-trivial extension to temporal label smoothing~\citet{yeche2023temporal} (TLS) for DSA.} (\textbf{iii}) At the event level, we propose a simple yet novel scheme, leveraging the risk localization provided by DSA models to prioritize imminent alarms, resulting in further performance improvement over EEP models.

\begin{figure*}[ht]
    \centering
    \includegraphics[width=\linewidth]{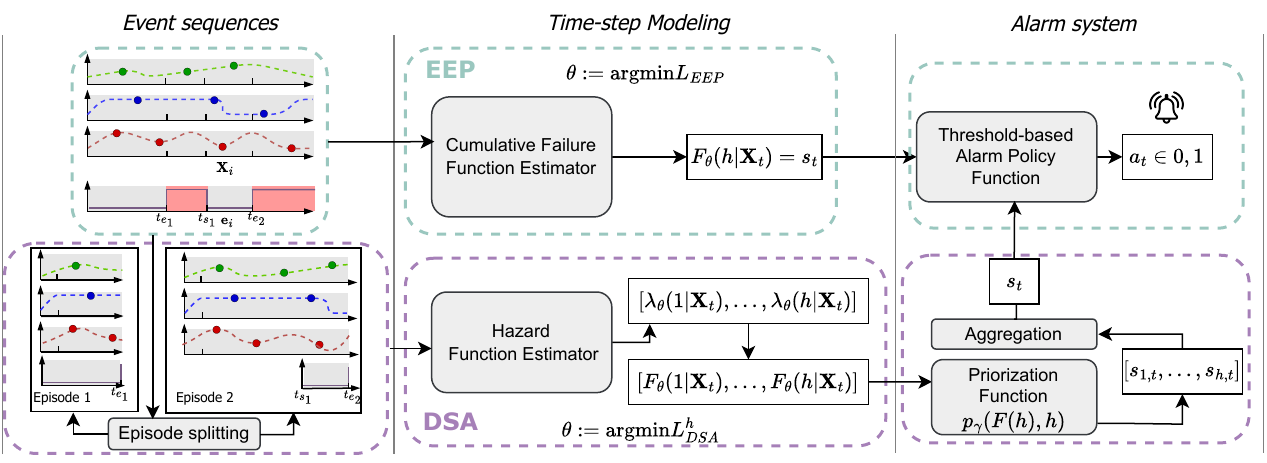}
    \caption{Overview of the pipeline for EEP (green) and our proposed DSA (purple) approach to EEP tasks. In the common EEP pipeline, a cumulative failure function estimator $F_\theta$ is trained by MLE on corresponding likelihood $L_{EEP}$ and serves as a risk score for a threshold-based alarm policy. On the other hand, in our proposed DSA approach, using a re-organized train set into episodes, we fit a hazard function estimator $\lambda_\theta$ by MLE on a partial survival likelihood $L^h_{DSA}$. We obtain a unique risk after applying a prioritization function and aggregation function to  $[F_\theta(1),...,F_\theta(h)]$.   }
    \label{fig:enter-label}
\end{figure*}

\section{Related Work}\label{sec:rw}

\paragraph{Early event prediction from EHR data} As previously mentioned, early warning systems (EWS) using deep learning models have recently gained traction in the literature. Indeed, over the years multiple large publicly available EHR databases~\citep{mimic-iii-johnson-nature16,pollard2018eicu,faltys2021hirid,thoral2021sharing} and benchmarks including EEP tasks~\citep{harutyunyan2019multitask,wang2020mimic,reyna2020early,yeche2021,vandewaterYetAnotherICUBenchmark2023} were released. Using these, existing works proposed new architecture designs~\citep{horn2020,tomavsev2019clinically}, imputation methods~\citep{futoma2017learning}, and more recently objective functions~\cite{yeche2023temporal}. However, in all these works, the backbone model is trained via MLE on the cumulative failure function at the horizon of prediction. Thus, to our knowledge, our work is the first to investigate the DSA models for these EEP tasks.

\paragraph{Survival analysis in the era of deep learning} 
With deep learning emergence, SA quickly moved away from proportional linear hazard models, as originally proposed by~\citet{cox1972regression}. A stream of work uses neural networks to parameterize the hazard function~\citep{yousefi2017predicting, katzman2018deepsurv}, \citet{gensheimer2019scalable} additionally remove the proportional hazard assumption. Simultaneously, other works focus on parameterizing the PMF~\citep{kvamme2019time,lee2018deephit, ren2019deep}, while adding regularization terms to their negative log-likelihood objective. Among these works, \citet{lee2018deephit}  went in the opposite direction to our work by fitting the PMF with MLE on the cumulative function, similar to EEP. In DSA, based on the landmarking idea~\citep{dsa-landmarking-houwelingen-2007, landmarking-sa-treatment-rct-2014}, advances followed a similar trend with works parametrizing the PMF~\citep{damera2022intervene} and fitting an MLE on the cumulative failure function~\citep{jarrett2019dynamic,lee2019dynamic}.

\paragraph{Survival analysis and event classification} \update{Prior work has investigated the use of static survival analysis for event classification problems such as early detection of fraud~\citep{surv-fraud-detection-aaai19-zheng}. This work however remains restricted to non-dynamic application, thus not applicable to DSA or EEP.
On the other hand, \citet{eeg-dsa-pmlr-shen23a} propose a model for EEP applications trained with a vanilla DSA likelihood to classify neurological prognostication. However, they do not provide any comparison to EEP MLE nor propose tailored alarm policies to the localized risk estimation. 
}

\section{Methods}
In this section, we describe our DSA approach to EEP tasks starting from the timestep estimator training to the alarm policy design. An overview of the pipeline and how it compares to EEP can be found in Figure~\ref{fig:enter-label}.
\subsection{From Early Event Prediction to Dynamic Survival Analysis}

\paragraph{Early event prediction} In EEP, we consider a dataset of multivariate time series of covariates $\mathbf{X}_{i}$ and binary event labels $e_{i,t}$ representing the occurrence of an event at time $t$ in trajectory $i$. Each sample $i$ is a sequence $\{ (\mathbf{x}_{i,0}, e_{i,0}), \ldots, (\mathbf{x}_{i,T_i}, e_{i,T_i}) \}$ of length $T_i$. For each timepoint $t$ along a time series, the covariates observed up to this point are denoted $\X_{i,t} = [\mathbf{x}_{i, 0}, \ldots,\mathbf{x}_{i,t}]$ and the time of the next event is given by $T_e(t) = \argmin_{\tau: \tau \geq t} \{e_{\tau}: e_{\tau} = 1 \}$. If no event happened, we define $T_e(t) = +\infty$ and call this sample "right-censored" to align with DSA terminology.  The EEP task consists of modeling the cumulative probability of this event occurring within a fixed prediction horizon $h$ defined as $F(h|\X_t) = P(T_e \leq t + h |\mathbf{X}_t)$. Importantly, as EEP focuses on early warning, no prediction is carried out during events. The common approach in the EEP literature is to train models that directly parameterize the cumulative failure function $F_\theta$ by minimizing the following negative log-likelihood with labels $y_{i,t} = \mathds{1}_{[\sum_{k=t}^{t+h} e_k \geq 1]}$:

\begin{multline}\label{eq:eep}
    L_{EEP} = \sum_i^N\sum_t^{T_i} \Bigl( -e_{i,t}[y_{i,t}\log(F_{\theta}(h|\X_{i,t})) \\ + (1-y_{i,t})\log(1-F_{\theta}(h|\X_{i,t}))] \Bigr)
\end{multline}

\paragraph{Dynamic survival analysis} Conversely, DSA aims to model the precise time $T_e(t)$ until an event of interest occurs given observation up to $t$. Thus, \update{when considering a discrete setting}, to model for all horizons $k\in \mathds{N}^*$, the mass function $f(k|\X_t) =  P(T_e = t + k |\mathbf{X}_t)$. This statistical framework has the particularity of considering terminal events only. There can be at most one event per trajectory $i$ and \update{if not censored}, it is at time $T_e = T_i +1 $. \update{Right-censoring refers to sequences where an event hasn't been observed before last observation $T_i$}. For this purpose, it is common to define $c_i = \mathds{1}_{[T_i \neq T_e]}$ a right-censoring indicator. Then the DSA negative log-likelihood for a model parameterizing the PMF $f_\theta$ is defined as: 

\begin{multline}
     L_{DSA} =  - \sum_{i=1}^N \sum_{t=0}^{T_i} \Bigl((1-c_i)f_\theta(T_i+1 -t | \mathbf{X}_{i,t})\\ + c_i([1-\sum_{k=1}^{T_i+1-t}f_\theta(k| \mathbf{X}_{i})]) \Bigr)
\end{multline}

It is known~\citep{kalbfleisch2011statistical} that the survival likelihood can be re-written as binary cross-entropy over the hazard function $\lambda(k|\X_t) =  P(T_e = t + k |\mathbf{X}_t, T_e > t + k - 1 )$. Thus, it is common in DSA to parameterize the hazard $\lambda_\theta$ and to minimize the following survival negative log-likelihood using binary labels $y_{i,t, k} = \mathds{1}_{[T_i - t = k \land c_i = 0]}$ \update{and sample weights $w_{i,t, k} = \mathds{1}_{[k \leq T_i - t ]}$} :

\begin{multline}\label{eq:dsa}
    L_{DSA} = \sum_{i=1}^N\sum_{t=0}^{T_i}\sum_{k=1}^{T_{\max}} w_{i,t, k}\\
    \Bigl( [y_{i,t,k}\log(\lambda_{\theta}(k|\X_{i,t})) \\ + (1-y_{i,t,k})\log(1-\lambda_{\theta}(k|\X_{i,t}))] \Bigr)
\end{multline}

Given an estimate $\lambda_\theta$, for a fixed $h$, we obtain estimates of the PMF $ f_\theta(h|\X_t) = \prod_{k=1}^{h-1}(1-\lambda_\theta(k|\X_t))\lambda_\theta(h|\X_t)$ and the cumulative failure function  $F_\theta(h|\X_t) = 1 - \prod_{k=1}^{h}(1-\lambda_\theta(k|\X_t))$. Thus, from a hazard estimate, one can also perform EEP tasks.

\paragraph{Handling non-terminal events}In general, events from EEP are not terminal, meaning that observations are carried out during and after an event and events can occur multiple times. To train a model using a survival analysis method for these cases, \update{the DSA framework requires unique terminal events. To address this issue, we propose to instead only predict the occurrence of the closest event, if there is one.

 In EEP tasks, timesteps within events are ignored in the likelihood. Thus, for a patient stay $i$ experiencing $v$ events at times $t_{e_1}, ..., t_{e_v}$, respectively ending at times $s_{e_1}, .. ., s_{e_v}$, we proposed instead to consider distinct episodes $[\mathbf{X}_{i,0},..., \mathbf{X}_{i,t_{e_1}-1}]$,$[\mathbf{X}_{i,0},..., \mathbf{X}_{i,t_{e_2}-1}]$, ..., $[\mathbf{X}_{i,0},..., \mathbf{X}_{i,T_i}]$ associated to their respective labels  $[\mathbf{y}_{i,0},..., \mathbf{y}_{i,t_{e_1}-1}]$,$[\mathbf{y}_{i,s_{e_2}},..., \mathbf{y}_{i,t_{e_2}-1}]$,...,$[\mathbf{y}_{i,s_{e_v}}$ $,...,\mathbf{y}_{i,T_i}]$. It is important to note that for any episode beyond the first one indexed by $k$, we provide the history of measurement between $0$ and $s_{e_{k-1}}$ to ensure preserving the signal from previous occurrences. This procedure ensures, that in each sample, if not censored, the event occurrence is unique and at the end of the sequence, as in DSA. }

\subsection{Bridging the gap on timestep-level performance}\label{sec:bridge} 

\begin{figure}[t]
    \centering
    \includegraphics[width=0.8\linewidth]{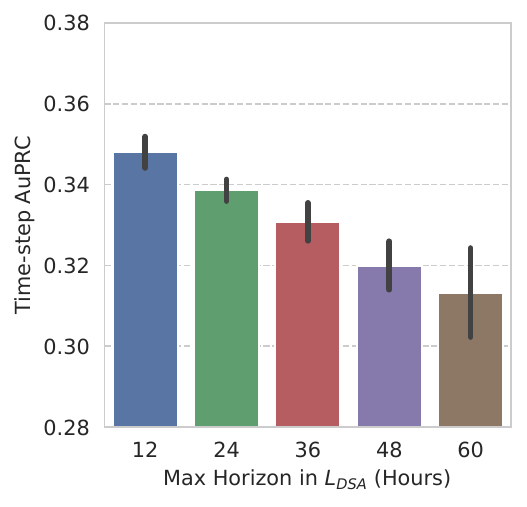}
    \caption{Ablation on the maximum considered horizon in $L_{DSA}$ for the ventilation task. By not training the DSA model further than $h$, the model's estimation of $F_\theta(h)$ improves leading to a better timestep AuPRC.}
    \label{fig:ab_vent}
\end{figure}

In practice,  \update{as reported by \citet{yeche2023temporal}, we show that fitting hazard deep learning models to a DSA likelihood on EEP data is unstable and underperforms on timestep metrics  (see Figure~\ref{fig:ab_vent} and Table~\ref{tab:timestep-benchmark-auprc})}. However,  we find we can overcome this issue with two simple modifications to the training. First, we find that instability is due to extreme imbalance compared to EEP likelihood and propose a specific \emph{logit bias initialization} to handle it. Second, to focus on the horizon of prediction used at inference, we propose to match EEP likelihood and \emph{truncate DSA likelihood only until the horizon of prediction $h$}. \update{Finally, we propose, survTLS our extension to TLS for survival analysis, allowing to further improve performance over DSA MLE for EEP.}

\begin{figure}[h]
    \centering
    \includegraphics[width=\linewidth]{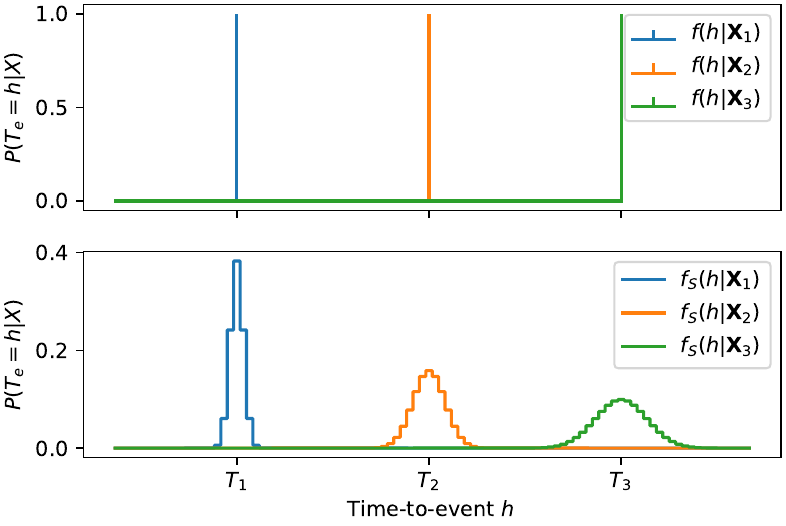}
    \caption{\update{Illustration of \textbf{survTLS} for three non-censored samples. (Top) Ground-truth PMF; (Bottom) survTLS smoothed PMF. The further the event, the higher the entropy of the mass function around the ground-truth time-to-event.}}
    \label{fig:survtls}
\end{figure}

\paragraph{Bias initialization}\label{sec:bias} As resolution is high in ICU data and horizons of prediction relatively short, associated tasks tend to already be imbalanced. Unfortunately, as shown in Eq~\ref{eq:dsa}, when fitting a hazard model, each positive label from the EEP is associated with $T_i -t - 1$ negative elements for a single positive label. Similarly, each negative is associated with $T_i -t$ negative labels. Hence the prevalence from the EEP task is divided by a factor $\Bar{T}$ corresponding to the average sequence length. This becomes extreme in EHR data where sequences have thousands of steps. We found this to forbid convergence in certain cases. 

To overcome this issue, inspired by \cite{karpathyblog}, we propose to initialize the bias of the logit layer $\mathbf{b} = [b_1, ..., b_{T_{\max}}]$  such that the output probability $\frac{1}{1+e^{-b_k}} = \Tilde{y}(k)$ with $\Tilde{y}(k)$, the average hazard label value for horizon $k$. Thus we initialize the bias as follows:
$$
b_k = \log(\frac{\Tilde{y}(k)}{1 - \Tilde{y}(k)}), \forall k \leq T_{\max} 
$$

\paragraph{Survival likelihood truncating }\label{sec:trunc}
As shown in Table~\ref{tab:timestep-benchmark-auprc}, correct initialization of biases already allows DSA models to be trainable for ICU data, however, they still lag behind EEP counterparts. As motivated by \cite{yeche2023temporal}, further events are generally harder to predict due to their lower signal. \update{We observe sequences to be longer in EEP datasets than in DSA. Indeed, PCB2 and AIDS, two commonly used datasets in DSA, have median lengths of 3 and 5 \cite{maystre2022temporally}, whereas MIMIC-III and HiRID have median sequence lengths of 50 and 275.} Thus, when fitting a DSA likelihood, a model is trained to model event occurrence possibly much further than the fixed horizon of prediction $h$ used in EEP. As empirically validated in Figure~\ref{fig:ab_vent}), we believe such events dominate the loss due to their hardness, forbidding the model to learn properly for events occurring within $h$ steps. In the alarm policy, no prediction, whether it is cumulative or not, is required beyond $h$.

 To solve this issue, because in EEP no prediction is carried beyond $h$, we propose to fit DSA models only until $h$. This translates into a truncated negative log-likelihood as follows:
\begin{multline}
    L^h_{DSA} = \sum_{i=1}^N\sum_{t=0}^{T_i}\sum_{k=1}^{h} w_{i,t,k} \\
        \Bigl( [y_{i,t,k}\log(\lambda_{\theta}(k|\X_{i,t})) \\
        + (1-y_{i,t,k})\log(1-\lambda_{\theta}(k|\X_{i,t}))] \Bigr)
\end{multline}

\update{
\paragraph{survTLS -- A temporal label smoothing approach for Survival Analysis} In EEP, prior work showed the effectiveness of TLS~\citep{yeche2023temporal} for timestep-level performance. As a form of regularization, this method enforces EEP model certainty on their estimate $F_\theta(h|\X_t)$ to decrease with the distance to the next event, by modulating similarly label smoothing strength during training. Given its success for EEP, transferring TLS to DSA is sensible. Unfortunately, this is not straightforward to do, as in DSA we model the more granular hazard function $\lambda_\theta$ over all horizons with the constraint that $\sum_h f_\theta(h) = 1$.

In our extension survTLS, we propose to leverage this higher granularity in labels, not to control the certainty of event occurrence, as in TLS, but rather to control the certainty of the event localization based on its distance.

For this purpose, as shown in Figure~\ref{fig:survtls}, we replace the (hard) ground truth PMF vector $\mathbf{f}_{i,t} = [f(h|\X_{i,t}) = \mathds{1}_{[T_i - t = h \land c_i = 0]}]_{h\geq1}$ by a smooth version $f_{S}$. Given a continuous distribution $ g_i \sim \mathcal{N}(T_i, (\frac{T_i}{l})^2) $ and $G_i$ its cumulative distribution, we define $f_S$ as its discretization : 

\begin{equation*}
f_{S}(h|\mathbf{X}_i)   = 
    \begin{cases}
    0  & \text{if }  c_i = 1 \\
    G_i(h + \frac{1}{2}) & \text{elif }  h = 1 \\
     G_i(h+\frac{1}{2}) \\\quad  - G_i(h - \frac{1}{2}) & \text{elif }  h \in [2, K -1] \\
    1 - G_i(h - \frac{1}{2}) & \text{elif }  h = T_{max} \\
    \end{cases}
\end{equation*}

The lengthscale hyperparameter $l$ control the strength of the smoothing. It was selected on validation metrics and more details can be found in Appendix~\ref{app:exp}. Note that we preserve $\sum_h f_{S}(h|\mathbf{X_{i,t}}) =  1$. Following discrete survival analysis definitions, we can define the smooth survival function $S_{S}(h|\mathbf{X}_{i,t}) = 1 - \sum_{k=1}^{h}f_S(h|\mathbf{X}_{i,t}) $  and the hazard function $\lambda_{S}(h|\mathbf{X}_{i,t}) = \frac{f_{S}(h|\mathbf{X}_{i,t})}{S_{S}(h- 1|\mathbf{X}_{i,t})}$. 
\vspace{1em}

By identifying $w_{ij}$ and $y_{ij}$ in Eq.~\ref{eq:dsa} as respectively the ground-truth survival $S(j|\mathbf{X})$ and hazard probability $\lambda(j|\mathbf{X})$ as proposed by \citet{maystre2022temporally}, we define our survTLS objective as follows:

\begin{multline}\label{eq:survtls}
    L_{survTLS} = \sum_{i=1}^N\sum_{t=0}^{T_i}\sum_{k=1}^{h} S_{S}(k|\X_{i,t})\\
    \Bigl( [\lambda_{S}(k|\X_{i,t}))\log(\lambda_{\theta}(k|\X_{i,t})) \\ + (1-\lambda_{S}(k|\X_{i,t})))\log(1-\lambda_{\theta}(k|\X_{i,t}))] \Bigr)
\end{multline}

We show the new labels $y$ and weights $w$ obtained from survTLS in Figure~\ref{fig:survtls_h} and Figure~\ref{fig:survtls_S} that can be found in Appendix~\ref{app:imp}

}

\subsection{Leveraging Hazard Estimation for Alarm Policy}

\label{sec:alarm-policy}

\paragraph{Threshold-based policy} In the literature, though crucial to clinical adoption, most works do not explore model performance in terms of alarms for whole events and rather focus on timestep modeling. This leads the current state-of-the-art alarm policy to still be straightforward. First, \citet{tomavsev2019clinically} proposed to define a working threshold $\tau \in [0,1]$ selected based on a timestep precision constraint. Then, the alarm policy raises alarm $a_t \in {0,1}$ at any timestep where risk score $s_t = F_\theta(h|\X_t)$ is above $\tau$ :

\begin{equation}\label{eq:alarm_base}
     a_t = \mathds{1}_{F_\theta(h|\X_t) \geq \tau}
\end{equation}

\paragraph{Silencing policy} Later, to reduce the false alarm rate, \citet{hyland2020} introduced the concept of silencing. After a raised alarm following Eq.~\ref{eq:alarm_base}. the system \emph{silences} all subsequent alarms until the duration $\sigma$ of the silencing time has passed. This was then adopted in subsequent works on EWS \citep{respiratory-hueser-2024,kidney-lyu-2024} If we define $d^a_t$ to be the distance to the last alarm at timestep $t$, the silenced alarm policy is defined as follows:

\begin{equation}\label{eq:alarm_silencing}
     a_t = \mathds{1}_{F_\theta(h|\X_t) \geq \tau}\mathds{1}_{d^a_t \geq \sigma}
\end{equation}

It is important to note that silencing is applied regardless of the correctness of the alarm. Indeed, EEP tasks are prognosis tasks, hence contrary to diagnosis tasks, the veracity of prediction is not verifiable until the event occurs.

A straightforward approach to using a DSA model is to extract the failure estimate $F_\theta(h|\X_t)$ and apply the same alarm policy. We refer to this approach as "Fixed horizon". However, our motivation to estimate the more challenging hazard function is to have as a counterpart an estimate on the localization of the risk within horizon $h$ for the alarm policy design. Given a risk estimate vector $\mathbf{r}_t = [F_\theta(1|\X_t),..., F_\theta(h|\X_t)]$, we formalize a mechanism to raise alarms depending on a unique functioning threshold compatible with the different event metric definitions.

\paragraph{Imminent prioritization policy} We follow the same intuition as~\citet{yeche2023temporal} to favor more imminent events given similar risk at horizon $h$. Indeed, impending events should be acted on immediately and are less likely to be impacted by a competitive event, thus they should have a higher priority than events equally probable but at a further horizon.

We formalize this intuition by introducing a priority function $p$ and transforming the output of the DSA model to create a score vector $\mathbf{s}_t \in [0,1]^h$ from the output vector $[[F_\theta(1|\X_t),..., F_\theta(h|\X_t)]]$ as follows:

\begin{align}
        \mathbf{s}_t &= [s_1, \dots,  s_h]  \nonumber \\
        &= [p(F_\theta(1|\X_t), 1),\dots, p(F_\theta(h|\X_t),h)]
\end{align}

\update{This step allows us to rescale risk predictions to a joint scale, from which we aggregate risk scores into a single alarm, using a single working threshold $\tau$, }by defining $a_t$ as follows:

\begin{equation}
    a_t = \mathds{1}_{(\sum \mathds{1}_{s_k > \tau})  > 0}\mathds{1}_{d^a_t \geq \sigma}
\end{equation}

Additionally we can define an estimate of the time-to-event $d_t = \min_k[k \hspace{0.5em} | \hspace{0.5em} s_k > \tau]$.

To enforce a prioritization of the closer horizons of prediction, we simply have to enforce, $p$ to be monotonically decreasing. We choose to implement the priority function with an exponential decay~\citep{yeche2023temporal}:

\begin{equation}
    p(F, k) = q^{exp}(k) \cdot F
\end{equation}

where the exponential decay function $q^{exp}(k)$ is defined as follows:

\begin{align}
    q(t) &= \begin{cases}
        0   & \text{if } k > h_{\max} \\
        e^{-\gamma (k-d)} + A  &  \text{if } k \leq h_{\max} \\
    \end{cases} \\
    \text{where}& \nonumber \\
    A &= -e^{-\gamma(h_{\max}-d)} \\
    d &= -\frac{1}{\gamma} ln(1 - e^{-\gamma h_{\max}})
\end{align}

\begin{figure}
    \centering
    \includegraphics[width=0.8\linewidth]{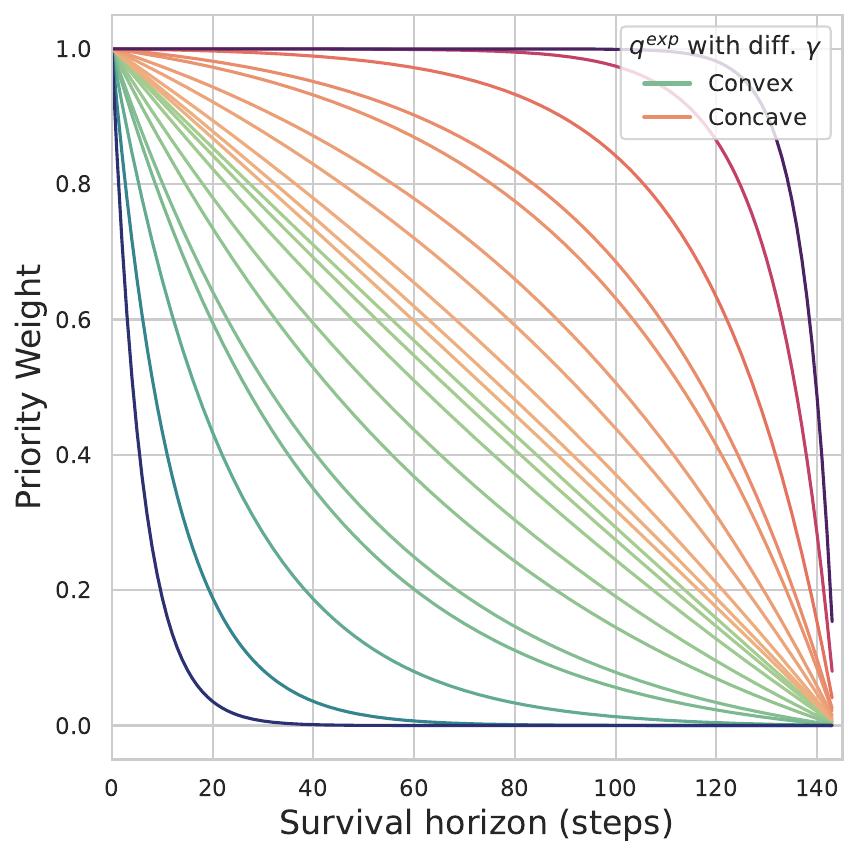}
    \caption{Visualizations of the $q_{exp}$ exponential decay function used for prioritization of survival risk scores. The plot shows $h_{\max}= h  =144$, and different $\gamma$ for the convex and a concave version, which we didn't find to work better.}
    \label{fig:q-exp-visualization}
\end{figure}

As shown in Figure~\ref{fig:q-exp-visualization}$, h_{\max}$ controls the intercept with 0 and $\gamma$ the strength of the decay. Hence, for any prediction beyond $h_{\max}$, the risk score is scaled to 0. The two hyperparameters $\gamma$ and $h_{\max}$ are tuned on the validation set Alarm/Event AuPRC and their value for different tasks is provided in Appendix~\ref{app:exp}.

\section{Experimental Set-up}

\paragraph{Tasks} We perform experiments on three EEP tasks, early prediction of \textit{circulatory failure}, \textit{mechanical ventilation} and \textit{decompensation} on established benchmarks from HiRID~\citep{yeche2021} and MIMIC-III~\citep{harutyunyan2019multitask}. Both circulatory failure and ventilation are predicted at a 12-hour horizon with a 5-minute resolution, while decompensation is predicted at 24 hours with a 1-hour resolution. Further details about tasks and dataset can be found in Appendix~\ref{app:data}

\paragraph{Implementation details} For all tasks, models are composed of a linear time-step embedding with $\mathcal{L}_1$-regularization~\citep{tomavsev2019clinically} coupled to a GRU~\citep{chung2014empirical} backbone. All hyperparameters shared across methods were selected through validation performance (AuPRC) for the EEP model and then used for all methods. Specific parameters to each method, such as priority strength, are selected on individual validation performance. Further details about implementation can be found in Appendix~\ref{app:imp}. As discussed in Section~\ref{sec:rw}, existing works have proposed specific improvements to either EEP likelihood, with auxiliary regression~\citep{tomavsev2019clinically} terms, or DSA likelihood with a ranking term~\citep{lee2018deephit,jarrett2019dynamic}. Both EEP and DSA likelihoods being versatile objectives, these extensions can be seamlessly incorporated for further applications. Hence, we focus on comparing directly likelihood objectives alone. We consider a model parameterizing the cumulative failure function at horizon $h$ fitted by MLE over the EEP likelihood and a model parameterizing the hazard function fitted by MLE over the DSA likelihood.

\paragraph{Evaluation} At a time-step level, to evaluate the goodness of the cumulative failure function estimate $F_\theta$ across models, we follow a common approach for highly imbalanced tasks, with the area under the precision-recall curve (AuPRC). We report the Alarm/Event AuPRC for event-level metrics, as defined by \cite{hyland2020}. To measure the timeliness of alarms, we also report the distance of the first alarm and event recall at high alarm precision thresholds. The high precision region represents a clinically relevant setting with a lower risk of alarm fatigue~\citep{tomavsev2019clinically,yeche2023temporal}. Unless stated otherwise, all results are reported in the form \textit{mean} $\pm$ \textit{95\% CI} of the standard error across 10 runs. In the following section, we describe the event-level metrics we used.

\subsection{Event-level Evaluation} \label{app:eval}
\citet{tomavsev2019clinically, yeche2023temporal, Moor2023-sepsis-lanzet} note the relevance of reporting event-oriented performance such as event recall or precision and distance to event at a fixed sensitivity level. To get a general event-level evaluation across different working thresholds, \citet{hyland2020} define event-based precision-recall curve mapping event recall to an effective alarm precision. We detail the event metric definitions below

\paragraph{Event Recall} Proposed by \citet{tomavsev2019clinically}, given binary timestep predictions $a_t$ (for alarms), event recall $R_{event}$ corresponds to the true positive rate over event predictions. An event $E$ at time $t_E$ is detected if any $a_t$ is positive between $t_E - h $ and $t_e-1$. Hence the following definition:
\begin{equation*}
    R_{event} = \frac{\sum_E \mathds{1}_{(\sum^{t_E-1}_{t_E-h}a_t)> 0}}{\#Events}
\end{equation*}

\paragraph{Alarm Precision} Proposed by \citet{hyland2020}, given binary timestep predictions $a_t$ (for alarms), alarm precision $P_{alarm}$ is the proportion of alarms raised located within $h$ of a event. It is defined as follows:
\begin{equation*}
    P_{event} = \frac{\sum_E(\sum^{t_E-1}_{t_E-h}a_t)}{\sum_t a_t}
\end{equation*}
Note, that considering a single event, there can be multiple true alarms for it as long as they fall within the prediction horizon.

In our work, the first event metric we report is the \textbf{Event Recall @ Alarm Precision} for high precision thresholds corresponding to clinically applicable regions. 

\paragraph{Distance to Event} Because event recall does not capture the earliness of the alarm, \citet{hyland2020} also proposes to report the distance for the first alarm for an event $D_{fa}$. Formally for an event event $E$ at time $t_E$ this is defined as:
\begin{equation*}
    D_{fa} = \max_{k \in [t_E-h, t_E-1]}[t_E - k | a_k = 1 ]
\end{equation*}

\paragraph{Alarm/Event AuPRC} Finally, to not depend on binary predictions, \citet{hyland2020} propose a curve score where alarms $\mathbf{a}$ directly depend on a working threshold $\tau \in [0, 1]$ where a point $(x, y) = \bigl( R_{event}(\tau), P_{alarm}(\tau) \bigr)$ on the curve. The final score is the area under this curve. As for timestep predictions, this metric gives a global overview of the model performance at different thresholds but on the granularity of events.

\section{Results}

\begin{table}[h]
    
    \caption{
        Time-Step AuPRC.\label{tab:timestep-benchmark-auprc} \update{ ``EEP" and ``Survival" refer to training with the respective vanilla likelihoods. Time-step level metrics are useful to assess a machine learning predictor's generalization performance on the trained task but do not explicitly quantify performance on clinically meaningful entities such as alarms and events.}
    }
    \footnotesize
    \setlength{\tabcolsep}{3pt}
\begin{tabular}{l c c c }
    \toprule
    \multirow{2}{*}{Task} & \multicolumn{2}{c}{HiRID} & MIMIC  \\
    \cmidrule(lr){2-2} \cmidrule(lr){3-4}
      & \textit{Circ.} & \textit{Vent.} & \textit{Decomp.} \\
    \midrule
    
\textbf{EEP} &  39.0 $\pm$ 0.4 &           34.3 $\pm$ 0.3 &    37.1 $\pm$ 0.6 \\

\update{+ TLS}  & \update{$\mathbf{40.5}$ $\pm$  $\mathbf{0.4}$} & \update{$\mathbf{34.9}$ $\pm$  $\mathbf{0.4}$ }     & \update{37.2 $\pm$ 0.3 } \\
    \midrule
\textbf{Survival} & 37.4 $\pm$ 2.0 &      21.5 $\pm$ 7.5 &  N.A \\
    \arrayrulecolor{lightgray}\midrule\arrayrulecolor{black}
    + Bias Init. & 39.2 $\pm$ 0.3 &       31.3 $\pm$ 0.8 &           36.2 $\pm$ 0.4 \\
    + Limit Horizon &  \textbf{40.7} $\pm$ 0.2 &             34.5 $\pm$ 0.4 &               37.4 $\pm$ 0.6\\
     \update{+ survTLS} & \update{$\mathbf{40.7}$ $\pm$ $\mathbf{0.2}$ }&  \update{$\mathbf{34.9}$ $\pm$ $\mathbf{0.3}$} & \update{$\mathbf{38.4}$ $\pm$ $\mathbf{0.2}$} \\

    \bottomrule
    \end{tabular}
\end{table}

\begin{table}[t]
    \centering
    \caption{\update{Area under the Alarm Precision / Event Recall Curve~\citep{hyland2020}. Different from time-step metrics (Table~\ref{tab:timestep-benchmark-auprc}), this metric evaluates the output of the alarm policy used over the model's time-step risk estimates. ``Survival" refers to training with bias initialization and truncated survival likelihood $L^h_{DSA}$. We observe that an alarm policy prioritizing imminent events can notably improve the event-level performance.}}
    \label{tab:event-auprc-benchmark}
    \footnotesize
\begin{tabular}{l c c c}
    \toprule
    \multirow{2}{*}{Task}  & \multicolumn{2}{c}{HiRID} & MIMIC \\
    \cmidrule(lr){2-3} \cmidrule(lr){4-4}
     & \textit{Circ.} & \textit{Vent.}   & \textit{Decomp.} \\
     \midrule
    
\textbf{EEP}  & 66.0$\pm$2.3 & 61.6$\pm$0.3 & 67.4$\pm$1.2 \\
    \arrayrulecolor{lightgray}\midrule\arrayrulecolor{black}
    \update{+ TLS} & \update{76.4$\pm$1.1} & \update{64.7$\pm$0.3} & \update{68.4$\pm$0.5} \\

\midrule

    \textbf{Survival} & 75.2$\pm$0.3 & 64.2$\pm$0.8  & 70.0$\pm$0.1 \\
    \arrayrulecolor{lightgray}\midrule\arrayrulecolor{black}

    \update{+ survTLS} & \update{77.5$\pm$0.6} & \update{64.8$\pm$1.5} & \update{70.9$\pm$0.3} \\
    
    \update{+ Imminent prio.} &  \update{\textbf{79.5$\pm$0.1}}  &  \update{\textbf{66.7$\pm$1.4}}   &  \update{\textbf{71.3$\pm$0.5	}} \\

    \bottomrule
    \end{tabular}
\end{table} 

\begin{table*}[t]
    \centering
    \captionof{table}{Event Recall and Mean Distance (in hours) of the first alarm to the event start on MIMIC-III Decompensation at fixed alarm precisions of 60\%, 70\%, 80\%. \update{In Table~\ref{tab:event-auprc-benchmark} we provide area under the curve results to assess the event-level performance across operating thresholds $\tau$. Here, to mimic deployment scenarios, we use a high precision constraint to fix a threshold on the risk score that has to be chosen for the alarm policy. Our method can provide noticeable improvements in event recall while maintaining a similar detection distance to the event.}}
\label{tab:event-benchmark-mimic}
    \begin{tabular}{l c c c c c c}
    \toprule
    
    \multirow{2}{*}{Metric} & \multicolumn{3}{c}{Event Recall} & \multicolumn{3}{c}{Mean Dist. (h)} \\
    \cmidrule(lr){2-4} \cmidrule(lr){5-7}
     & \textit{ @ 60\% P.} & \textit{@ 70\% P.} & \textit{@ 80\% P.} & \textit{@ 60\% P.} & \textit{@ 70\% P.} & \textit{@ 80\% P.} \\ \midrule
    
EEP & 66.7$\pm$0.0 & 60.9$\pm$0.1 & 52.4$\pm$0.0 & 6.2$\pm$0.1 & 4.9$\pm$0.1 & 3.4$\pm$0.1 \\
    \arrayrulecolor{lightgray}\midrule\arrayrulecolor{black}
    \update{+ TLS}  &  \update{68.1$\pm$0.0}&	\update{62.4$\pm$0.0} &	\update{54.8$\pm$0.0} &	\update{6.5$\pm$0.0} &	  \update{\textbf{5.2$\pm$0.0}} &	\update{ 3.7$\pm$0.0 }\\

\midrule
\textbf{Survival}  & 68.5$\pm$0.0 & 63.2$\pm$0.0  & 57.5$\pm$0.0	  & 6.4$\pm$0.1  &  5.1$\pm$0.1	 & \textbf{4.0$\pm$0.1} \\
    \arrayrulecolor{lightgray}\midrule\arrayrulecolor{black}

    \update{+ survTLS} & \update{68.8$\pm$0.1} &	\update{64.1$\pm$0.1} &	\update{58.5$\pm$0.0} &	\update{6.4$\pm$0.1} &	\update{\textbf{5.2$\pm$0.1}} &	\update{\textbf{4.0$\pm$0.1}} \\
    \update{+ Imminent prio.} & \update{\textbf{70.3$\pm$0.1}} &	\update{\textbf{64.6$\pm$0.0}} &	\update{\textbf{59.9$\pm$0.0}} &	\update{\textbf{6.6$\pm$0.1}} &	\update{5.1$\pm$0.1} &	\update{\textbf{4.0$\pm$0.1}} \\

    \bottomrule
    \end{tabular}

 \end{table*}
\paragraph{Time-step level} As shown in Table~\ref{tab:timestep-benchmark-auprc}, we find that vanilla survival models are significantly worse than their EEP counterparts going as far as not converging on the decompensation task. However, fixing bias initialization, as proposed in Section~\ref{sec:bias}, allows convergence for all tasks. Additionally, by truncating the DSA likelihood up to $h$ (Section~\ref{sec:trunc}), we managed to close the gap and even surpass in some cases MLE with EEP likelihood at a time-step level.

\begin{figure}[t]
    \centering
    \includegraphics[trim={6cm 0 6cm 0},clip,width=0.95\linewidth]{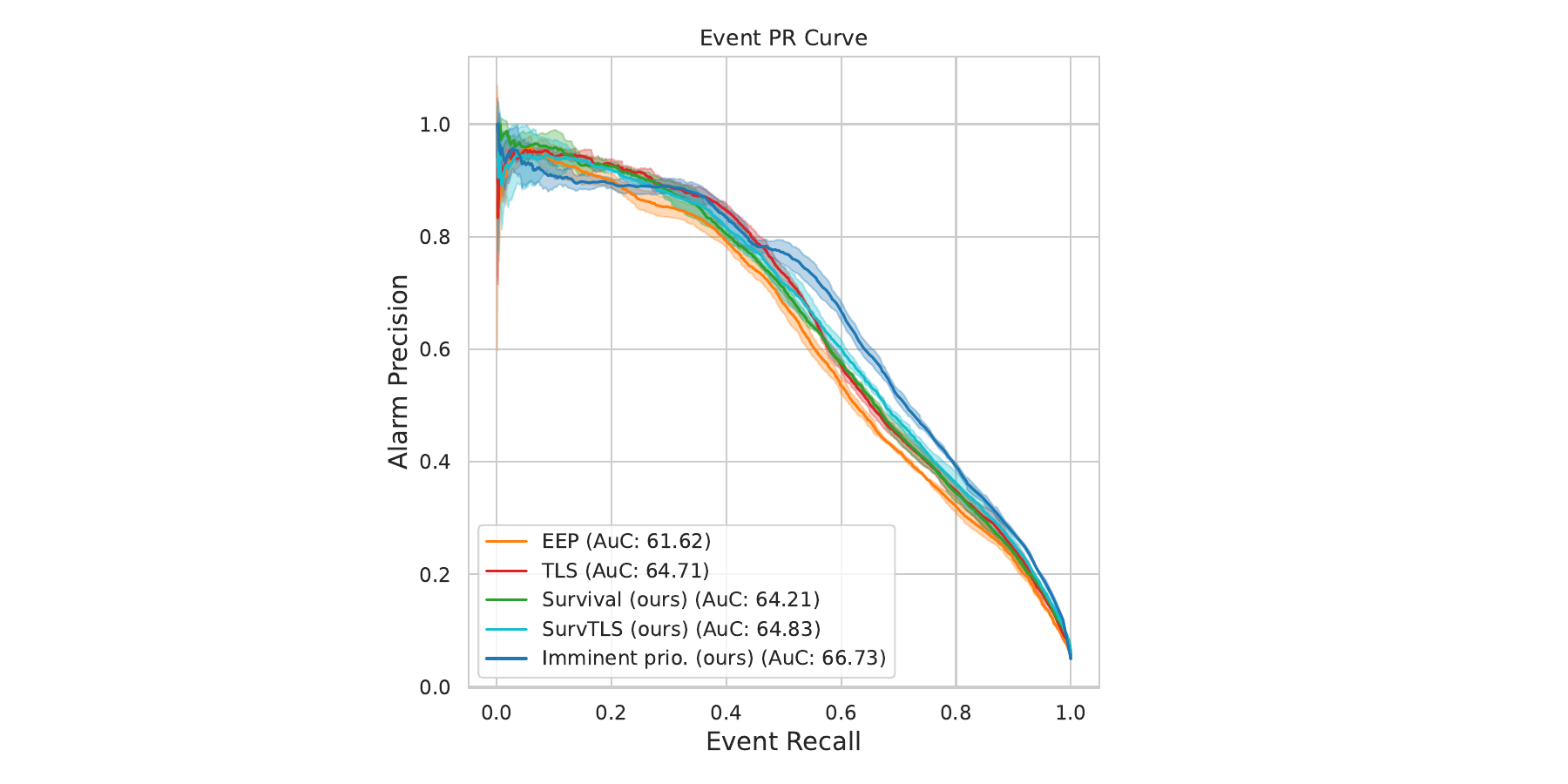}
    \caption{Alarm Precision~\citep{hyland2020} per Event Recall on HiRID Ventilation. \update{We show that our proposed DSA-based approach paired with a prioritizing alarm policy maintains a higher alarm precision for higher event recall regions.}}
    \label{fig:vent-event-precision}
\end{figure}

\begin{figure}[t]
    \centering
    \includegraphics[trim={6cm 0 6cm 0},clip,width=0.95\linewidth]{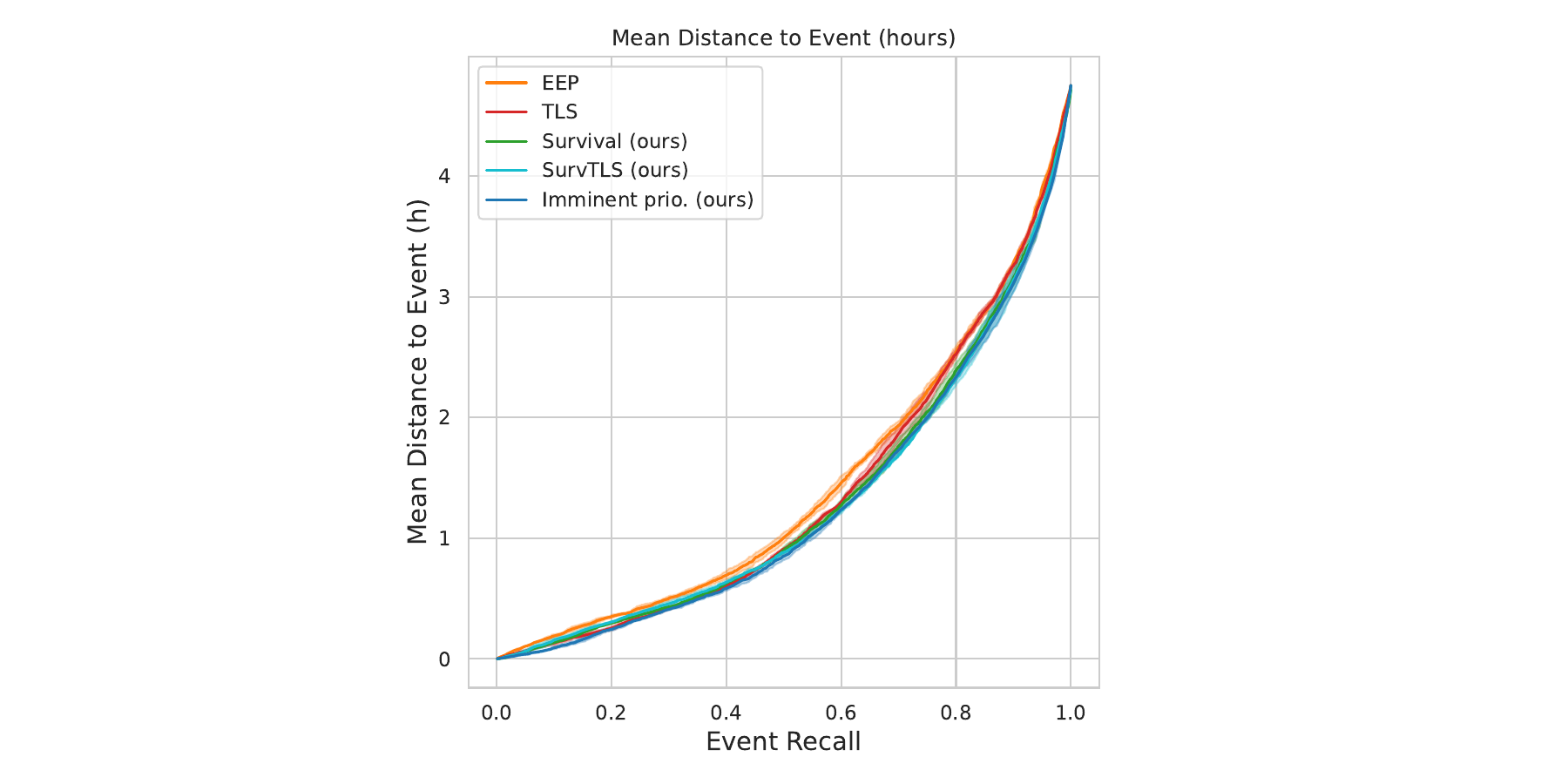}
    \caption{Mean distance of the first alarm to the event per Event Recall on HiRID Ventilation. \update{It highlights that the improved event recall and alarm precision performance does come at a very small to no cost in terms of detection distance to the event.}}
    \label{fig:vent-event-mean-distance}
\end{figure}

\paragraph{Event level} As shown in Table~\ref{tab:event-auprc-benchmark}, we find that survival models, given the prior improvements, already outperform EEP when used with a similar alarm policy with a fixed threshold over $F_\theta(h)$. When additionally introducing a priority function favoring imminent events, this gap further increases by 1 to 3\%. Finally, as shown in Table~\ref{tab:event-benchmark-mimic} for decompensation, our prioritization of more imminent events does not come at the cost of event recall nor distance to the event as our policy still matches or outperforms the base policy in both metrics. Similar conclusions can be drawn for HiRID tasks (Appendix~\ref{app:results}).

Finally, in Figures~\ref{fig:vent-event-precision} and~\ref{fig:vent-event-mean-distance} we show the full performance curve over all operating thresholds $\tau \in [0,1]$. The curves show alarm precision and mean distance of the first alarm over event recall on the HiRID Ventilation task, the two other benchmarked tasks are shown in Appendix~\ref{app:results}.

\section{Limitation and Further Work}

The focus of our study is to compare two distinct, yet related, modeling approaches on their ability to predict the likelihood of a future event. In an isolated comparison of the probabilistic modeling approach, our research establishes that by training a richer DSA model, one can successfully perform established EEP tasks. However, it also has some clear limitations. As mentioned in Section~\ref{sec:rw}, multiple works have proposed to complement both objectives with regularization terms. If interested in absolute performance for a specific application, one should ensure that DSA improvements remain compatible with such extensions. Also, DSA likelihood suffers from a much higher class imbalance that grows with resolution. While we manage to tackle this issue in our tasks of interest, it is still likely that for other EEP tasks, this instability might remain.
\update{Also note that, as we focused on EEP applications, the notion of competitive risk is not considered (a setting where we aim to predict different types of events with a joint modeling approach). Extending to multiple risks constitutes a non-trivial, but highly relevant future work.}

Furthermore, we highlight how \update{the richer DSA} output can be leveraged to enhance clinically relevant and deployment-oriented event-level performance metrics by prioritizing imminent events. However, our approach is straightforward and remains a first step. We hypothesize that this richer output can be further utilized to create more sophisticated alarm policies, which can be tuned to be more patient-specific by incorporating the shape of the individual estimated survival distributions at a given time point for a given patient. Additionally, this approach can provide clinicians with richer insights and explanations. We reserve this direction for further work.

\section{Conclusion}

In this work, we investigate the usage of DSA models for EEP tasks motivated by the additional localization of the risk they provide. We show that even though it is more challenging to train, with careful initialization and partial survival likelihood fitting, DSA models can be competitive at a time-step level. Further, we show that our simple prioritization scheme for alarms allows DSA models to outperform EEP counterparts even further. Our proposed prioritization transformation is a first step towards tailored alarm policies. Future work remains to further leverage risk localization to design more sophisticated alarm policies and provide a richer output to clinicians.

In the past many application-oriented works have chosen to use classical EEP modeling approaches for their studies due to higher stability and ease of training. We hope our analysis paves the way for DSA, a more comprehensive modeling approach, to become the prominent choice.

\clearpage
\bibliography{chil-sample}

\begin{thebibliography}{41}
\providecommand{\natexlab}[1]{#1}
\providecommand{\url}[1]{\texttt{#1}}
\expandafter\ifx\csname urlstyle\endcsname\relax
  \providecommand{\doi}[1]{doi: #1}\else
  \providecommand{\doi}{doi: \begingroup \urlstyle{rm}\Url}\fi

\bibitem[Chung et~al.(2014)Chung, Gulcehre, Cho, and
  Bengio]{chung2014empirical}
Junyoung Chung, Caglar Gulcehre, KyungHyun Cho, and Yoshua Bengio.
\newblock Empirical evaluation of gated recurrent neural networks on sequence
  modeling.
\newblock \emph{arXiv preprint arXiv:1412.3555}, 2014.

\bibitem[Cox(1972)]{cox1972regression}
David~R Cox.
\newblock Regression models and life-tables.
\newblock \emph{Journal of the Royal Statistical Society: Series B
  (Methodological)}, 34\penalty0 (2):\penalty0 187--202, 1972.

\bibitem[Damera~Venkata and Bhattacharyya(2022)]{damera2022intervene}
Niranjan Damera~Venkata and Chiranjib Bhattacharyya.
\newblock When to intervene: Learning optimal intervention policies for
  critical events.
\newblock \emph{Advances in Neural Information Processing Systems},
  35:\penalty0 30114--30126, 2022.

\bibitem[Di~Giuseppe et~al.(2016)Di~Giuseppe, Pappenberger, Wetterhall,
  Krzeminski, Camia, Libert{\'a}, and San~Miguel]{di2016potential}
Francesca Di~Giuseppe, Florian Pappenberger, Fredrik Wetterhall, Blazej
  Krzeminski, Andrea Camia, Giorgio Libert{\'a}, and Jesus San~Miguel.
\newblock The potential predictability of fire danger provided by numerical
  weather prediction.
\newblock \emph{Journal of Applied Meteorology and Climatology}, 55\penalty0
  (11):\penalty0 2469--2491, 2016.

\bibitem[Faltys et~al.(2021{\natexlab{a}})Faltys, Zimmermann, Lyu, Hüser,
  Hyland, Rätsch, and Merz]{hirid-faltys-physionet21}
M.~Faltys, M.~Zimmermann, X.~Lyu, M.~Hüser, S.~Hyland, G.~Rätsch, and
  T.~Merz.
\newblock Hirid, a high time-resolution icu dataset, 2021{\natexlab{a}}.

\bibitem[Faltys et~al.(2021{\natexlab{b}})Faltys, Zimmermann, Lyu, H{\"u}ser,
  Hyland, R{\"a}tsch, and Merz]{faltys2021hirid}
Martin Faltys, Marc Zimmermann, Xinrui Lyu, Matthias H{\"u}ser, Stephanie
  Hyland, Gunnar R{\"a}tsch, and Tobias Merz.
\newblock Hirid, a high time-resolution icu dataset (version 1.1. 1).
\newblock \emph{Physio. Net}, 10, 2021{\natexlab{b}}.

\bibitem[Futoma et~al.(2017)Futoma, Hariharan, and Heller]{futoma2017learning}
Joseph Futoma, Sanjay Hariharan, and Katherine Heller.
\newblock Learning to detect sepsis with a multitask gaussian process rnn
  classifier.
\newblock In \emph{International conference on machine learning}, pages
  1174--1182. PMLR, 2017.

\bibitem[Gensheimer and Narasimhan(2019)]{gensheimer2019scalable}
Michael~F Gensheimer and Balasubramanian Narasimhan.
\newblock A scalable discrete-time survival model for neural networks.
\newblock \emph{PeerJ}, 7:\penalty0 e6257, 2019.

\bibitem[Harutyunyan et~al.(2019)Harutyunyan, Khachatrian, Kale, Ver~Steeg, and
  Galstyan]{harutyunyan2019multitask}
Hrayr Harutyunyan, Hrant Khachatrian, David~C Kale, Greg Ver~Steeg, and Aram
  Galstyan.
\newblock Multitask learning and benchmarking with clinical time series data.
\newblock \emph{Scientific data}, 6\penalty0 (1):\penalty0 1--18, 2019.

\bibitem[Horn et~al.(2020)Horn, Moor, Bock, Rieck, and Borgwardt]{horn2020}
Max Horn, Michael Moor, Christian Bock, Bastian Rieck, and Karsten Borgwardt.
\newblock Set functions for time series.
\newblock In \emph{International Conference on Machine Learning}, pages
  4353--4363. PMLR, 2020.

\bibitem[Hyland et~al.(2020)Hyland, Faltys, H{\"{u}}ser, Lyu, Gumbsch, Esteban,
  Bock, Horn, Moor, Rieck, et~al.]{hyland2020}
Stephanie~L Hyland, Martin Faltys, Matthias H{\"{u}}ser, Xinrui Lyu, Thomas
  Gumbsch, Crist{\'{o}}bal Esteban, Christian Bock, Max Horn, Michael Moor,
  Bastian Rieck, et~al.
\newblock Early prediction of circulatory failure in the intensive care unit
  using machine learning.
\newblock \emph{Nature medicine}, 26\penalty0 (3):\penalty0 364--373, 2020.

\bibitem[Hüser et~al.(2024)Hüser, Lyu, Faltys, Pace, Hoche, Hyland,
  Y{\`e}che, Burger, Merz, and Rätsch]{respiratory-hueser-2024}
Matthias Hüser, Xinrui Lyu, Martin Faltys, Aliz{\'e}e Pace, Marine Hoche,
  Stephanie Hyland, Hugo Y{\`e}che, Manuel Burger, Tobias~M Merz, and Gunnar
  Rätsch.
\newblock A comprehensive ml-based respiratory monitoring system for
  physiological monitoring \& resource planning in the icu.
\newblock \emph{medRxiv}, 2024.
\newblock \doi{10.1101/2024.01.23.24301516}.
\newblock URL
  \url{https://www.medrxiv.org/content/early/2024/01/23/2024.01.23.24301516}.

\bibitem[Jarrett et~al.(2019)Jarrett, Yoon, and van~der
  Schaar]{jarrett2019dynamic}
Daniel Jarrett, Jinsung Yoon, and Mihaela van~der Schaar.
\newblock Dynamic prediction in clinical survival analysis using temporal
  convolutional networks.
\newblock \emph{IEEE journal of biomedical and health informatics}, 24\penalty0
  (2):\penalty0 424--436, 2019.

\bibitem[Johnson et~al.(2016{\natexlab{a}})Johnson, Pollard, and
  Mark]{mimic-iii-johnson-physionet16}
Alistair E.~W. Johnson, Tom~J. Pollard, and Roger~G. Mark.
\newblock {MIMIC-III} clinical database (version 1.4), 2016{\natexlab{a}}.

\bibitem[Johnson et~al.(2016{\natexlab{b}})Johnson, Pollard, Shen, Lehman,
  Feng, Ghassemi, Moody, Szolovits, Anthony~Celi, and
  Mark]{mimic-iii-johnson-nature16}
Alistair~E.W. Johnson, Tom~J. Pollard, Lu~Shen, Li-wei~H. Lehman, Mengling
  Feng, Mohammad Ghassemi, Benjamin Moody, Peter Szolovits, Leo Anthony~Celi,
  and Roger~G. Mark.
\newblock Mimic-iii, a freely accessible critical care database.
\newblock \emph{Scientific Data}, 3\penalty0 (1):\penalty0 160035, May
  2016{\natexlab{b}}.
\newblock ISSN 2052-4463.
\newblock \doi{10.1038/sdata.2016.35}.
\newblock URL \url{https://doi.org/10.1038/sdata.2016.35}.

\bibitem[Kalbfleisch and Prentice(2011)]{kalbfleisch2011statistical}
John~D Kalbfleisch and Ross~L Prentice.
\newblock \emph{The statistical analysis of failure time data}.
\newblock John Wiley \& Sons, 2011.

\bibitem[Karpathy(2019)]{karpathyblog}
Andrej Karpathy.
\newblock A recipe for training neural networks.
\newblock \emph{Personal Blog}, 2019.
\newblock URL \url{https://karpathy.github.io/2019/04/25/recipe/}.

\bibitem[Katzman et~al.(2018)Katzman, Shaham, Cloninger, Bates, Jiang, and
  Kluger]{katzman2018deepsurv}
Jared~L Katzman, Uri Shaham, Alexander Cloninger, Jonathan Bates, Tingting
  Jiang, and Yuval Kluger.
\newblock Deepsurv: personalized treatment recommender system using a cox
  proportional hazards deep neural network.
\newblock \emph{BMC medical research methodology}, 18\penalty0 (1):\penalty0
  1--12, 2018.

\bibitem[Kingma and Ba(2017)]{kingma2017adam}
Diederik~P. Kingma and Jimmy Ba.
\newblock Adam: A method for stochastic optimization, 2017.

\bibitem[Kvamme et~al.(2019)Kvamme, Borgan, and Scheel]{kvamme2019time}
H{\aa}vard Kvamme, {\O}rnulf Borgan, and Ida Scheel.
\newblock Time-to-event prediction with neural networks and cox regression.
\newblock \emph{arXiv preprint arXiv:1907.00825}, 2019.

\bibitem[Lee et~al.(2018)Lee, Zame, Yoon, and Van Der~Schaar]{lee2018deephit}
Changhee Lee, William Zame, Jinsung Yoon, and Mihaela Van Der~Schaar.
\newblock Deephit: A deep learning approach to survival analysis with competing
  risks.
\newblock In \emph{Proceedings of the AAAI conference on artificial
  intelligence}, volume~32, 2018.

\bibitem[Lee et~al.(2019)Lee, Yoon, and Van Der~Schaar]{lee2019dynamic}
Changhee Lee, Jinsung Yoon, and Mihaela Van Der~Schaar.
\newblock Dynamic-deephit: A deep learning approach for dynamic survival
  analysis with competing risks based on longitudinal data.
\newblock \emph{IEEE Transactions on Biomedical Engineering}, 67\penalty0
  (1):\penalty0 122--133, 2019.

\bibitem[Lyu et~al.(2024)Lyu, Fan, H{\"u}ser, Hartout, Gumbsch, Faltys, Merz,
  R{\"a}tsch, and Borgwardt]{kidney-lyu-2024}
Xinrui Lyu, Bowen Fan, Matthias H{\"u}ser, Philip Hartout, Thomas Gumbsch,
  Martin Faltys, Tobias~M. Merz, Gunnar R{\"a}tsch, and Karsten Borgwardt.
\newblock An empirical study on kdigo-defined acute kidney injury prediction in
  the intensive care unit.
\newblock \emph{medRxiv}, 2024.
\newblock \doi{10.1101/2024.02.01.24302063}.
\newblock URL
  \url{https://www.medrxiv.org/content/early/2024/02/03/2024.02.01.24302063}.

\bibitem[Maystre and Russo(2022)]{maystre2022temporally}
Lucas Maystre and Daniel Russo.
\newblock Temporally-consistent survival analysis.
\newblock \emph{Advances in Neural Information Processing Systems},
  35:\penalty0 10671--10683, 2022.

\bibitem[Moor et~al.(2023)Moor, Bennett, Ple{\v c}ko, Horn, Rieck, Meinshausen,
  B{\"u}hlmann, and Borgwardt]{Moor2023-sepsis-lanzet}
Michael Moor, Nicolas Bennett, Drago Ple{\v c}ko, Max Horn, Bastian Rieck,
  Nicolai Meinshausen, Peter B{\"u}hlmann, and Karsten Borgwardt.
\newblock Predicting sepsis using deep learning across international sites: a
  retrospective development and validation study.
\newblock \emph{EClinicalMedicine}, 62:\penalty0 102124, August 2023.

\bibitem[Parast et~al.(2014)Parast, Tian, and
  Cai]{landmarking-sa-treatment-rct-2014}
Layla Parast, Lu~Tian, and Tianxi Cai.
\newblock Landmark estimation of survival and treatment effect in a randomized
  clinical trial.
\newblock \emph{J Am Stat Assoc}, 109\penalty0 (505):\penalty0 384--394,
  January 2014.

\bibitem[Pollard et~al.(2018)Pollard, Johnson, Raffa, Celi, Mark, and
  Badawi]{pollard2018eicu}
Tom~J Pollard, Alistair~EW Johnson, Jesse~D Raffa, Leo~A Celi, Roger~G Mark,
  and Omar Badawi.
\newblock The eicu collaborative research database, a freely available
  multi-center database for critical care research.
\newblock \emph{Scientific data}, 5\penalty0 (1):\penalty0 1--13, 2018.

\bibitem[Ren et~al.(2019)Ren, Qin, Zheng, Yang, Zhang, Qiu, and
  Yu]{ren2019deep}
Kan Ren, Jiarui Qin, Lei Zheng, Zhengyu Yang, Weinan Zhang, Lin Qiu, and Yong
  Yu.
\newblock Deep recurrent survival analysis.
\newblock In \emph{Proceedings of the AAAI Conference on Artificial
  Intelligence}, volume~33, pages 4798--4805, 2019.

\bibitem[Reyna et~al.(2020)Reyna, Josef, Jeter, Shashikumar, Westover, Nemati,
  Clifford, and Sharma]{reyna2020early}
Matthew~A Reyna, Christopher~S Josef, Russell Jeter, Supreeth~P Shashikumar,
  M~Brandon Westover, Shamim Nemati, Gari~D Clifford, and Ashish Sharma.
\newblock Early prediction of sepsis from clinical data: the
  physionet/computing in cardiology challenge 2019.
\newblock \emph{Critical care medicine}, 48\penalty0 (2):\penalty0 210--217,
  2020.

\bibitem[Shen et~al.(2023)Shen, Elmer, and Chen]{eeg-dsa-pmlr-shen23a}
Xiaobin Shen, Jonathan Elmer, and George~H. Chen.
\newblock Neurological prognostication of post-cardiac-arrest coma patients
  using eeg data: A dynamic survival analysis framework with competing risks.
\newblock In Kaivalya Deshpande, Madalina Fiterau, Shalmali Joshi, Zachary
  Lipton, Rajesh Ranganath, Iñigo Urteaga, and Serene Yeung, editors,
  \emph{Proceedings of the 8th Machine Learning for Healthcare Conference},
  volume 219 of \emph{Proceedings of Machine Learning Research}, pages
  667--690. PMLR, 11--12 Aug 2023.
\newblock URL \url{https://proceedings.mlr.press/v219/shen23a.html}.

\bibitem[Sutton et~al.(2020)Sutton, Pincock, Baumgart, Sadowski, Fedorak, and
  Kroeker]{sutton2020overview}
Reed~T Sutton, David Pincock, Daniel~C Baumgart, Daniel~C Sadowski, Richard~N
  Fedorak, and Karen~I Kroeker.
\newblock An overview of clinical decision support systems: benefits, risks,
  and strategies for success.
\newblock \emph{NPJ digital medicine}, 3\penalty0 (1):\penalty0 17, 2020.

\bibitem[Thoral et~al.(2021)Thoral, Peppink, Driessen, Sijbrands, Kompanje,
  Kaplan, Bailey, Kesecioglu, Cecconi, Churpek, et~al.]{thoral2021sharing}
Patrick~J Thoral, Jan~M Peppink, Ronald~H Driessen, Eric~JG Sijbrands, Erwin~JO
  Kompanje, Lewis Kaplan, Heatherlee Bailey, Jozef Kesecioglu, Maurizio
  Cecconi, Matthew Churpek, et~al.
\newblock Sharing icu patient data responsibly under the society of critical
  care medicine/european society of intensive care medicine joint data science
  collaboration: the amsterdam university medical centers database
  (amsterdamumcdb) example.
\newblock \emph{Critical care medicine}, 49\penalty0 (6):\penalty0 e563, 2021.

\bibitem[Toma{\v{s}}ev et~al.(2019)Toma{\v{s}}ev, Glorot, Rae, Zielinski,
  Askham, Saraiva, Mottram, Meyer, Ravuri, Protsyuk,
  et~al.]{tomavsev2019clinically}
Nenad Toma{\v{s}}ev, Xavier Glorot, Jack~W Rae, Michal Zielinski, Harry Askham,
  Andre Saraiva, Anne Mottram, Clemens Meyer, Suman Ravuri, Ivan Protsyuk,
  et~al.
\newblock A clinically applicable approach to continuous prediction of future
  acute kidney injury.
\newblock \emph{Nature}, 572\penalty0 (7767):\penalty0 116--119, 2019.

\bibitem[Tutz et~al.(2016)Tutz, Schmid, et~al.]{tutz2016modeling}
Gerhard Tutz, Matthias Schmid, et~al.
\newblock \emph{Modeling discrete time-to-event data}.
\newblock Springer, 2016.

\bibitem[van~de Water et~al.(2024)van~de Water, Schmidt, Elbers, Thoral,
  Arnrich, and Rockenschaub]{vandewaterYetAnotherICUBenchmark2023}
Robin van~de Water, Hendrik Schmidt, Paul Elbers, Patrick Thoral, Bert Arnrich,
  and Patrick Rockenschaub.
\newblock Yet another {ICU} benchmark: A flexible multi-center framework for
  clinical {ML}.
\newblock In \emph{The Twelfth International Conference on Learning
  Representations}, 2024.
\newblock URL \url{https://openreview.net/forum?id=ox2ATRM90I}.

\bibitem[Van~Houwelingen(2007)]{dsa-landmarking-houwelingen-2007}
Hans~C. Van~Houwelingen.
\newblock Dynamic prediction by landmarking in event history analysis.
\newblock \emph{Scandinavian Journal of Statistics}, 34\penalty0 (1):\penalty0
  70--85, 2007.
\newblock \doi{https://doi.org/10.1111/j.1467-9469.2006.00529.x}.
\newblock URL
  \url{https://onlinelibrary.wiley.com/doi/abs/10.1111/j.1467-9469.2006.00529.x}.

\bibitem[Wang et~al.(2020)Wang, McDermott, Chauhan, Ghassemi, Hughes, and
  Naumann]{wang2020mimic}
Shirly Wang, Matthew~BA McDermott, Geeticka Chauhan, Marzyeh Ghassemi,
  Michael~C Hughes, and Tristan Naumann.
\newblock Mimic-extract: A data extraction, preprocessing, and representation
  pipeline for mimic-iii.
\newblock In \emph{Proceedings of the ACM conference on health, inference, and
  learning}, pages 222--235, 2020.

\bibitem[Y{\`{e}}che et~al.(2021)Y{\`{e}}che, Kuznetsova, Zimmermann,
  H{\"{u}}ser, Lyu, Faltys, and R{\"{a}}tsch]{yeche2021}
Hugo Y{\`{e}}che, Rita Kuznetsova, Marc Zimmermann, Matthias H{\"{u}}ser,
  Xinrui Lyu, Martin Faltys, and Gunnar R{\"{a}}tsch.
\newblock Hirid-icu-benchmark--a comprehensive machine learning benchmark on
  high-resolution icu data.
\newblock \emph{arXiv preprint arXiv:2111.08536}, 2021.

\bibitem[Y{\`e}che et~al.(2023)Y{\`e}che, Pace, Ratsch, and
  Kuznetsova]{yeche2023temporal}
Hugo Y{\`e}che, Aliz{\'e}e Pace, Gunnar Ratsch, and Rita Kuznetsova.
\newblock Temporal label smoothing for early event prediction.
\newblock \emph{ICML}, 2023.

\bibitem[Yousefi et~al.(2017)Yousefi, Amrollahi, Amgad, Dong, Lewis, Song,
  Gutman, Halani, Velazquez~Vega, Brat, et~al.]{yousefi2017predicting}
Safoora Yousefi, Fatemeh Amrollahi, Mohamed Amgad, Chengliang Dong, Joshua~E
  Lewis, Congzheng Song, David~A Gutman, Sameer~H Halani, Jose~Enrique
  Velazquez~Vega, Daniel~J Brat, et~al.
\newblock Predicting clinical outcomes from large scale cancer genomic profiles
  with deep survival models.
\newblock \emph{Scientific reports}, 7\penalty0 (1):\penalty0 11707, 2017.

\bibitem[Zheng et~al.(2019)Zheng, Yuan, and
  Wu]{surv-fraud-detection-aaai19-zheng}
Panpan Zheng, Shuhan Yuan, and Xintao Wu.
\newblock Safe: A neural survival analysis model for fraud early detection.
\newblock \emph{Proceedings of the AAAI Conference on Artificial Intelligence},
  33\penalty0 (01):\penalty0 1278--1285, Jul. 2019.
\newblock \doi{10.1609/aaai.v33i01.33011278}.
\newblock URL \url{https://ojs.aaai.org/index.php/AAAI/article/view/3923}.

\end{thebibliography}

\clearpage
\appendix
\onecolumn

\section{Experimental details} \label{app:exp}

\subsection{Datasets} \label{app:data}

\begin{table*}[h]
    \centering
    \caption{\textit{Label and event prevalence statistics} computed on the training set for all tasks. }
    \label{tab:event_prevalence}
\begin{tabular}{lccc}
    \toprule
        \multirow{2}{*}{Task} & Positive & Patients undergoing  & Number of events \\
        &  timesteps (\%) & event (\%) & per positive patient\\\midrule
        Circulatory Failure (HiRID) & 4.3 & 25.6 & 1.9 \\
        Mechanical Ventilation (HiRID) & 5.6 & 56.5 & 1.5 \\
        Decompensation (MIMIC) & 2.1 & 8.3 & 1.0\\
        \bottomrule
    \end{tabular}\end{table*}
\paragraph{Circulatory failure}

Online binary prediction of future circulatory failure events on the HiRID~\citep{faltys2021hirid} dataset as defined by~\citet{yeche2021} every 5 minutes. The benchmarked prediction horizon for the EEP models and the survival model at a fixed horizon are set at 12 hours.

\paragraph{Ventilation}

Online binary prediction of future ventilator usage on the HiRID~\citep{faltys2021hirid} dataset every 5 minutes. The ventilation status is extracted from the data as defined by~\citet{yeche2021} and the prediction horizon is 12 hours.

\paragraph{Decompensation}

Online binary prediction of patient mortality as defined by~\citet{harutyunyan2019multitask}. A label is positive if the patient dies within the horizon. Benchmarked and evaluated on MIMIC-III~\citep{mimic-iii-johnson-nature16} at a 24-hour horizon.

\subsection{Implementation} \label{app:imp}

\paragraph{Training details.} For all models, we set the batch size to 64 and the learning rate to $1e^{-4}$ using Adam optimizer~\cite{kingma2017adam}. We early-stop each model training according to their validation loss when no improvement was made after 10 epochs. 

\paragraph{Libraries.} An exhaustive list of libraries and their version we used is in the \texttt{environment.yml} file from the code release.

\paragraph{Infrastructure.}
We trained all models on a single \texttt{NVIDIA RTX2080Ti} with 8 \texttt{Xeon E5-2630v4} cores and 64GB of memory. Individual seed training took between 3 to 10 hours for each run.

\paragraph{Timestep modeling hyperparameters} We used the same architecture and shared hyperparameters for both types of likelihood training. These were selected based on the validation performance of  EEP likelihood training. It is possible further improvement can be achieved by selecting different hyperparameters for our DSA approach. However, we prefer to be conservative to ensure a fair comparison. Exact parameters are reported in Table~\ref{tab:hp-search-circ}, Table~\ref{tab:hp-search-vent}, Table~\ref{tab:hp-search-decomp}.

\begin{figure*}[t]
    \centering
    \includegraphics[width=\linewidth]{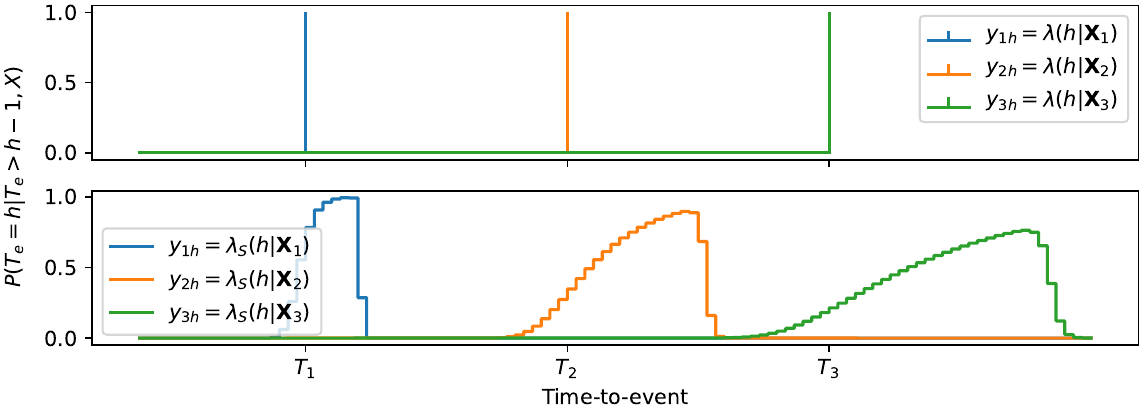}
    \caption{\update{Illustration of the smooth hazard function $\lambda_S$ serving as labels in \textbf{survTLS}.}}
    \label{fig:survtls_h}
\end{figure*}

\begin{figure*}[h]
    \centering
    \includegraphics[width=\linewidth]{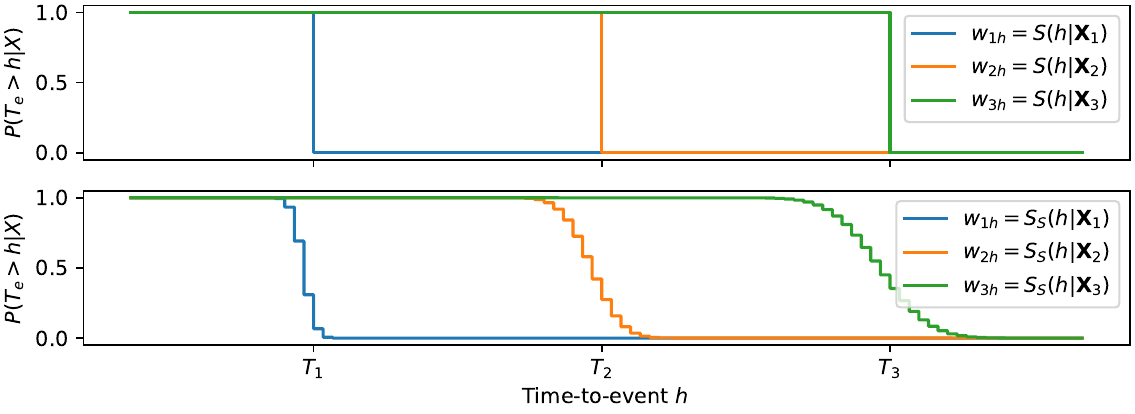}
    \caption{\update{Illustration of the smooth survival function $\S_S$ serving as weights in \textbf{survTLS}.}}
    \label{fig:survtls_S}
\end{figure*}

\paragraph{Temporal label smoothing hyperparameters} Similarly we selected hyperparameters for TLS and survTLS on validation set timestep AUPRC. We found similar hyperparameters as the original paper for TLS with $h_max = 2h$ and $h_min=0$ and $\gamma_{circ} = 0.2$, $\gamma_{vent} = 0.1$, and $\gamma_{decomp} = 0.05$. For survTLS, we found for the lengthscale parameter that $l_{circ}=10$, $l_{vent}=50$, {$l_{decomp}=8$}. We plot the obtained labels and weights from smoothing the groud-truth event PMF $f$  corresponding to smooth hazard function $\lambda_S$ and survival function $S_S$ in Figure~\ref{fig:survtls_h} and  Figure~\ref{fig:survtls_S}.

\paragraph{Alarm policy hyperparameter} We find that a short one-step silencing actually performs the best (5 minutes on HiRID and 1 hour on MIMIC-III) based on validation set performance for the EEP model and keep that constant across all experiments also for the survival models.

For the prioritized alarm policy we tune $h_{max}$, $\gamma$, and $q^{exp}$ function type for each task on the validation set. Chosen values are shown in Table~\ref{tab:hp-search-prio}.

\begin{table*}[tbh!]
    \centering
\caption{Hyperparameter search range for prioritized alarm policies on top of DSA models. In \textbf{bold} are parameters selected by grid search.}
\begin{tabular}{l c c c}
\toprule
Hyperparameter & Decomp & Circ. & Vent. \\
\midrule
\midrule
\midrule
$h_{max}$ & (\textbf{12}, 24, 36, 48, 96) & \textbf{144}:[72,720] & \textbf{576}:[72,720] \\
\midrule
$\gamma$ & \textbf{0.1}:[0.01,2.0] & \textbf{0.5}:[0.01,2.0] & \textbf{2.0}:[0.01,2.0] \\
\midrule
Function Type & (\textbf{Convex}, Concave) & (\textbf{Convex}, Concave) & (\textbf{Convex}, Concave) \\
\bottomrule
\end{tabular}
\label{tab:hp-search-prio}
\end{table*}

\begin{table}[tbh!]
    \centering
\caption{Hyperparameter search range for \textit{circulatory failure}, In \textbf{bold} are parameters selected by grid search.}
\begin{tabular}{lc}
\toprule
Hyperparameter & Values\\
\midrule
\midrule
\midrule
Drop-out & (\textbf{0.0}, 0.1, 0.2, 0.3) \\
\midrule
Depth &   (1, \textbf{2}, 3, 4) \\
\midrule
Hidden Dimension & (64, 128, \textbf{256}, 512) \\
\midrule
L1 Regularization &  (1e-1, 1, \textbf{10}, 100)\\
\bottomrule
\end{tabular}
\label{tab:hp-search-circ}
\end{table}

\begin{table}[tbh!]
    \centering
\caption{Hyperparameter search range for \textit{mechanical ventilation}, In \textbf{bold} are parameters selected by grid search.}
\begin{tabular}{lc}
\toprule
Hyperparameter & Values\\
\midrule
\midrule
\midrule
Drop-out & (\textbf{0.0}, 0.1, 0.2, 0.3) \\
\midrule
Depth &   (1, 2, \textbf{3}, 4) \\
\midrule
Hidden Dimension & (128, 256, \textbf{512},1024) \\
\midrule
L1 Regularization &  (1e-1, 1, \textbf{10}, 100)\\
\bottomrule
\end{tabular}
\label{tab:hp-search-vent}
\end{table}

\begin{table}[tbh!]
    \centering
\caption{Hyperparameter search range for \textit{decompensation}, In \textbf{bold} are parameters selected by grid search.}
\begin{tabular}{lc}
\toprule
Hyperparameter & Values\\
\midrule
\midrule
\midrule
Drop-out & (0.0, 0.1, \textbf{0.2}, 0.3) \\
\midrule
Depth &   ( 2, 3, \textbf{4}, 5) \\
\midrule
Hidden Dimension & (128, \textbf{256}, 512 ,1024) \\
\midrule
L1 Regularization &  (1e-1, \textbf{1}, 10, 100)\\
\bottomrule
\end{tabular}
\label{tab:hp-search-decomp}
\end{table}

\section{Additional Results} \label{app:results}

\begin{table*}
    \centering
    \caption{Event Performance on HiRID Circulatory Failure at fixed alarm precisions.}
    \label{tab:event-benchmark-circ}
\begin{tabular}{l c c c c c c}
    \toprule
    
    \multirow{2}{*}{Metric} & \multicolumn{3}{c}{Event Recall} & \multicolumn{3}{c}{Mean Dist. (h)} \\
    \cmidrule(lr){2-4} \cmidrule(lr){5-7}
     & \textit{@ 60\% P.} & \textit{@ 70\% P.} & \textit{@ 80\% P.} & \textit{@ 60\% P.} & \textit{@ 70\% P.} & \textit{@ 80\% P.} \\ \midrule
    
EEP & 67.1$\pm$0.0 & 46.6$\pm$0.0 & 24.5$\pm$0.1 & 1.61$\pm$0.04 & 0.93$\pm$0.02 & 0.36$\pm$0.02 \\
    \arrayrulecolor{lightgray}\midrule\arrayrulecolor{black}
    + TLS  & 83.3$\pm$0.0 &	70.9$\pm$0.0 &	52.8$\pm$0.0 &	\textbf{1.99$\pm$0.04} &	\textbf{1.40$\pm$0.01} &	\textbf{0.81$\pm$0.02} \\
    
\midrule

    \textbf{Survival}  & 79.0$\pm$0.0 & 65.0$\pm$0.0 & 48.2$\pm$0.0 & 1.81$\pm$0.02 & 1.19$\pm$0.02 & 0.69$\pm$0.01 \\
    \arrayrulecolor{lightgray}\midrule\arrayrulecolor{black}

    \update{+ survTLS} & \update{82.5$\pm$0.0} &	\update{69.5$\pm$0.0} &\update{54.4$\pm$0.0} &	\update{1.85$\pm$0.02} &	\update{1.26$\pm$0.02} &	\update{0.78$\pm$0.02} \\
    \update{+ Imminent prio.} & \update{\textbf{88.7$\pm$0.0}} &	\update{\textbf{75.0$\pm$0.0}} &	\update{\textbf{55.3$\pm$0.0}} &	\update{1.91$\pm$0.03} &	\update{1.24$\pm$0.04} &	\update{0.70$\pm$0.00} \\

    \bottomrule
    \end{tabular}
\end{table*} \begin{table*}
    \centering
    \caption{Event Performance on HiRID Ventilation at fixed alarm precisions.}
    \label{tab:event-benchmark-vent}
\begin{tabular}{l c c c c c c}
    \toprule
    
    \multirow{2}{*}{Metric} & \multicolumn{3}{c}{Event Recall} & \multicolumn{3}{c}{Mean Dist. (h)} \\
    \cmidrule(lr){2-4} \cmidrule(lr){5-7}
     & \textit{ @ 60\% P.} & \textit{ @ 70\% P.} & \textit{ @ 80\% P.} & \textit{ @ 60\% P.} & \textit{@ 70\% P.} & \textit{@ 80\% P.} \\ \midrule
    
EEP & 55.7$\pm$0.0 & 49.0$\pm$0.0 & 39.7$\pm$0.0 & 1.25$\pm$0.04 & 0.96$\pm$0.04 & 0.69$\pm$0.03 \\

    \arrayrulecolor{lightgray}\midrule\arrayrulecolor{black}
    + TLS & 58.2$\pm$0.1 &	52.9$\pm$0.0 &	44.7$\pm$0.0 &	1.23$\pm$0.01 &	1.01$\pm$0.00 &	\textbf{0.73$\pm$0.01} \\

\midrule
\textbf{Survival} & 58.2$\pm$0.0 & 50.8$\pm$0.0 & 41.2$\pm$0.0 & 1.20$\pm$0.03 & 0.93$\pm$0.03 & 0.66$\pm$0.01 \\
    \arrayrulecolor{lightgray}\midrule\arrayrulecolor{black}

    \update{+ survTLS} & \update{60.3$\pm$0.0} &	\update{52.3$\pm$0.0} &	\update{42.7$\pm$0.0} &	\update{1.24$\pm$0.02} &	\update{0.94$\pm$0.01} &	\update{0.69$\pm$0.01} \\
    \update{+ Imminent prio.} & \update{\textbf{64.5$\pm$0.1}} &	\update{\textbf{58.0$\pm$0.1}} &	\update{\textbf{43.8$\pm$0.0}} &	\update{\textbf{1.43$\pm$0.02}} &	\update{\textbf{1.14$\pm$0.03}} &	\update{0.66$\pm$0.03} \\

    \bottomrule
    \end{tabular}
    
\end{table*} 
\paragraph{Event Recall and Distance to Event} In Tables~\ref{tab:event-benchmark-circ} and~\ref{tab:event-benchmark-vent} we show event recall and mean distance to events at different alarm precision levels for HiRID Ventilation and Circulatory Failure respectively. As noted before in the main manuscript, also on the HiRID dataset we can improve event performance while maintaining (or even slightly improving) on the distance of the first alarm to the event.

\begin{figure*}[htbp]
    \centering
    \subfigure[Decompensation on MIMIC-III]{
        \includegraphics[width=0.7\linewidth]{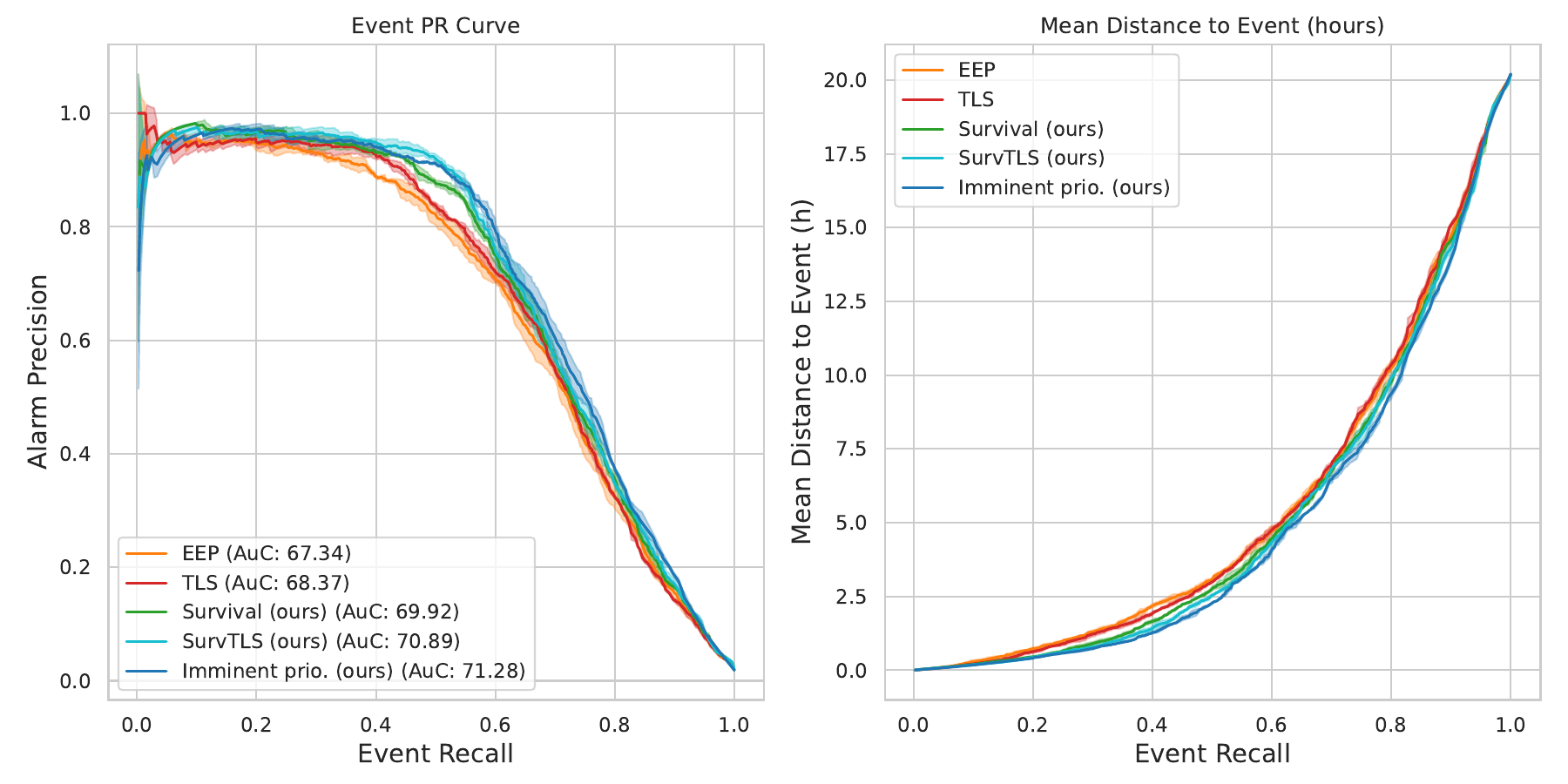}
    }
    \subfigure[Circulatory Failure on HiRID]{
        \includegraphics[width=0.7\linewidth]{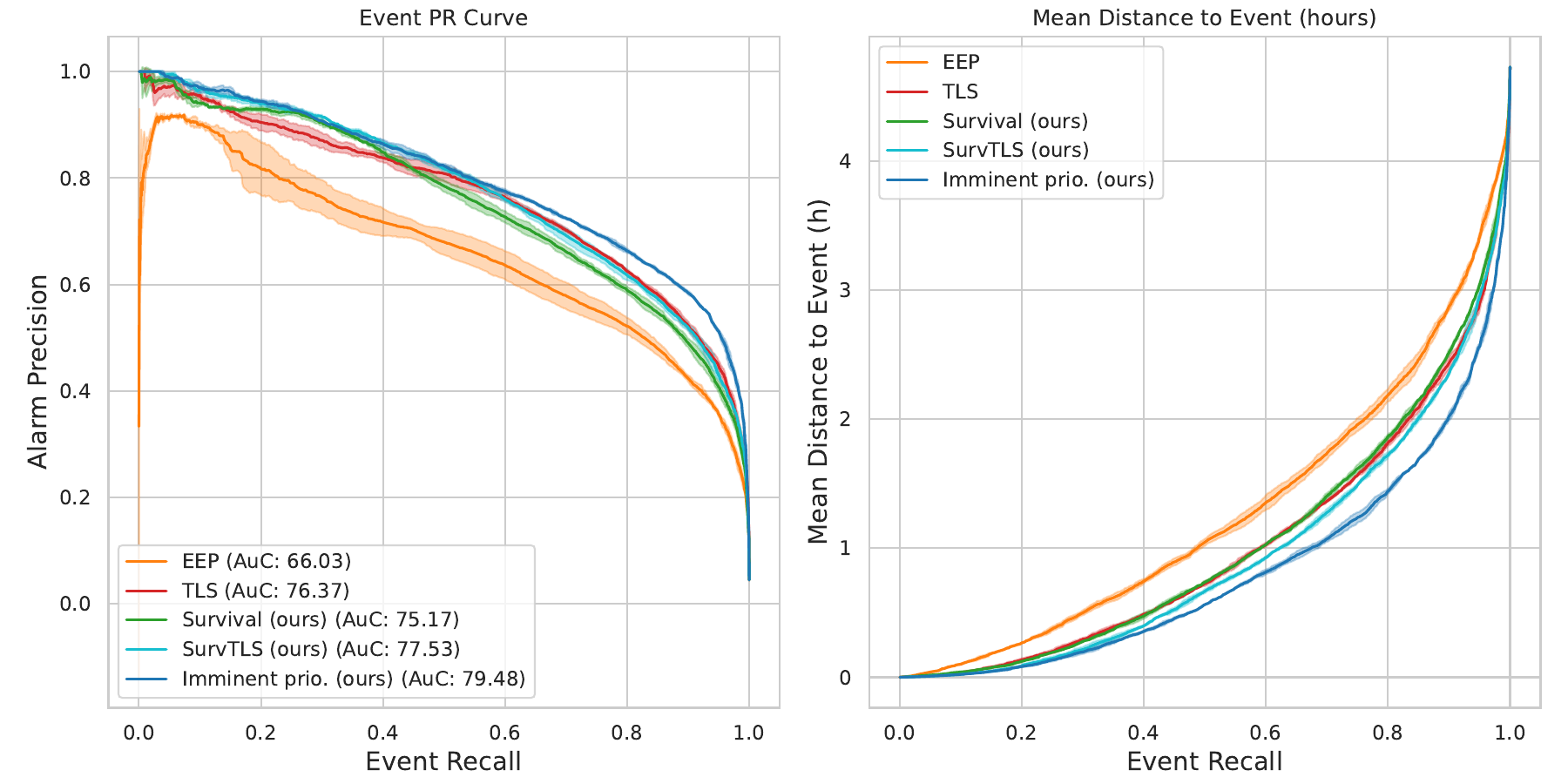}
    }
    \subfigure[Ventilation on HiRID]{
        \includegraphics[width=0.7\linewidth]{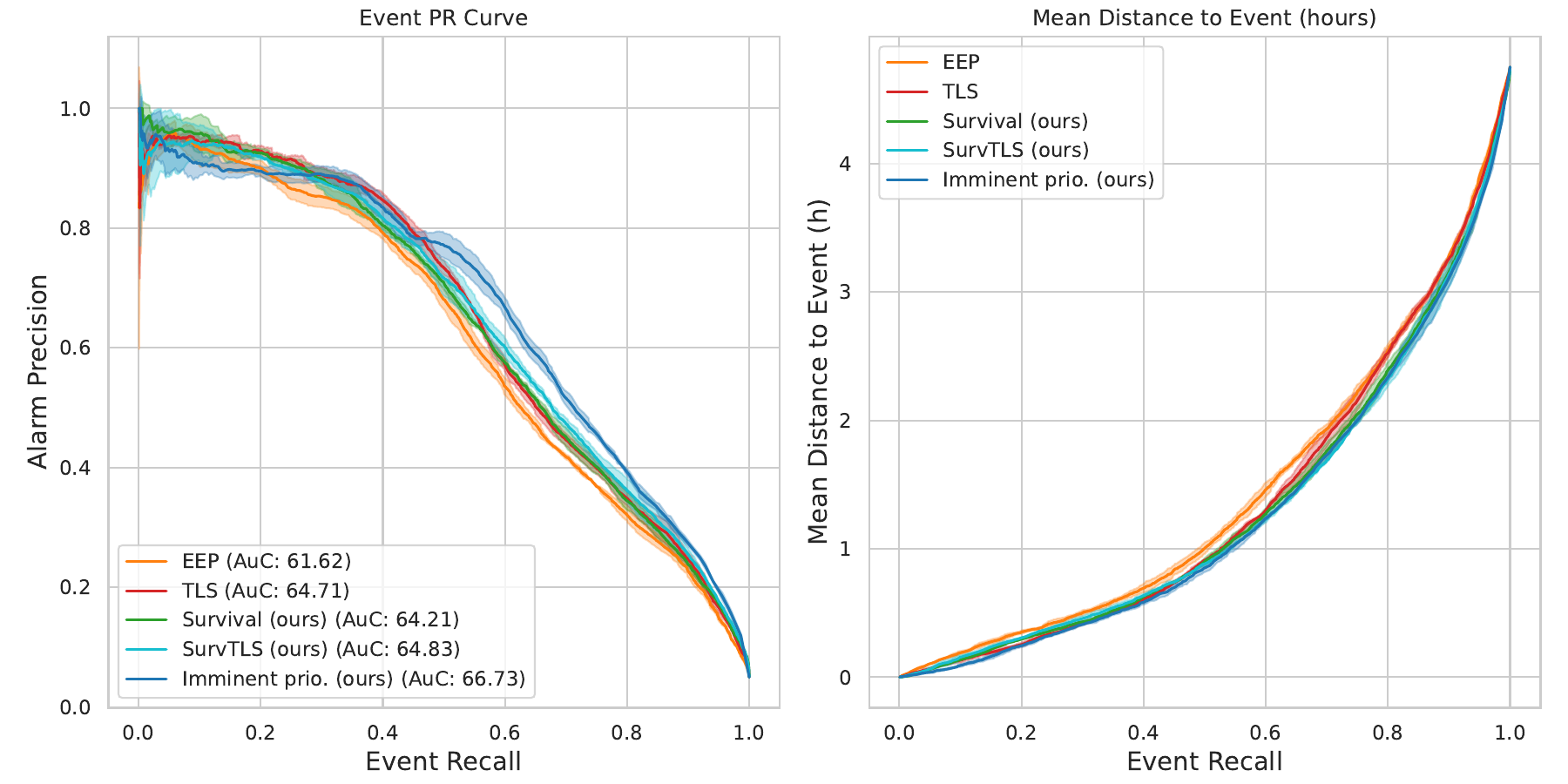}
    }
    \caption{Event-based performances of the alarm policy under different models. We show the event-based alarm precision~\citep{hyland2020} and also plot the mean distance of the first alarm to the event horizon at each event detection sensitivity level.}
    \label{fig:event-alarm-curves}
\end{figure*}

\paragraph{Event Performance Curves} In Figure~\ref{fig:event-alarm-curves} we show alarm precision and mean distance to event curves plotted over levels of event recall (sensitivity of the alarm policy conditioned on a risk predictor).

\end{document}